\documentclass[journal]{IEEEtran}

\usepackage[T1]{fontenc}
\usepackage[utf8]{inputenc}
\usepackage{amsmath}
\usepackage{amssymb}
\usepackage{bm}
\usepackage{mathtools}
\usepackage{graphicx}
\usepackage{booktabs}
\usepackage{tabularx}
\usepackage{multirow}
\usepackage{subcaption}
\usepackage{siunitx}
\usepackage{algorithm}
\usepackage{algpseudocode}
\usepackage{cite}
\usepackage{xcolor}
\usepackage{colortbl}
\usepackage{nicematrix}
\definecolor{measc}{HTML}{CFE2F3}\definecolor{predc}{HTML}{FCE5CD}
\usepackage{tikz}
\usetikzlibrary{positioning,arrows.meta,fit,backgrounds,calc}
\usepackage[hidelinks]{hyperref}
\usepackage{cleveref}
\crefname{figure}{Fig.}{Figs.}\Crefname{figure}{Fig.}{Figs.}\crefname{section}{Section}{Sections}\Crefname{section}{Section}{Sections}
\crefname{table}{Table}{Tables}\Crefname{table}{Table}{Tables}
\crefformat{equation}{#2(#1)#3}
\Crefformat{equation}{Equation~#2(#1)#3}

\newcommand{\vect}[1]{\mathbf{#1}}
\newcommand{\mat}[1]{\mathbf{#1}}

\newcommand{\meas}[1]{\tilde{#1}}
\newcommand{\real}{\mathbb{R}}
\newcommand{\SE}[1]{\mathrm{SE}(#1)}

\newcommand{\Log}{\mathrm{Log}}
\newcommand{\brina}{\textsc{Brina}}

\begin{document}

\title{\brina: Bottom-Referenced Inertial Navigation with AKAZE, a Sonar SLAM for Autonomous Marine Vehicles}

\newif\ifanonymous
\ifdefined\isanonymous\anonymoustrue\fi
\ifanonymous
\author{Anonymous Authors}
\markboth{IEEE Transactions on Robotics}{Seabed-edge sonar--inertial SLAM}
\else
\author{Adolfo~Damiano~Cafaro, Patrizio~Mariani, and~Roberto~Galeazzi
\thanks{A.~D.~Cafaro and R.~Galeazzi are with DTU Electro, and P.~Mariani is with DTU Aqua, Technical
University of Denmark, 2800 Kongens Lyngby, Denmark (e-mail: \{adaca, roga\}@dtu.dk, pat@aqua.dtu.dk).}}
\markboth{IEEE Transactions on Robotics}{Cafaro \MakeLowercase{\textit{et al.}}: Seabed-edge sonar--inertial SLAM}
\fi

\maketitle

\begin{abstract}
Forward-looking sonar (FLS) provides reliable underwater perception where optical sensing is limited, but its use for navigation is hindered by the unobserved elevation of each return. As a result, motion estimated directly from sonar images suffers from an unknown metric scale that is commonly resolved using a Doppler velocity log (DVL). We present \brina{} (Bottom-Referenced Inertial Navigation with AKAZE), a sonar--inertial SLAM system for marine vehicles that uses the FLS as its only exteroceptive sensor. \brina{} estimates the sonar altitude from the first seabed return in each beam and uses it to project matched features onto the seabed, providing metric odometry without a DVL. We further derive a bound relating the remaining scale error to seabed tilt and slope variability. The method is evaluated in simulation with an autonomous underwater vehicle, and in field experiments with uncrewed surface vehicles over ten missions, using two FLS systems under different operating conditions.
\end{abstract}


\section{Introduction}
Marine vehicles surveying the seabed and inspecting critical infrastructure require reliable localisation when external positioning is unavailable or untrusted. For uncrewed surface vehicles (USVs), GNSS can be degraded by multipath and shadowing near harbour structures, or falsified by spoofing~\cite{makar2023gnss,bhatti2017spoofing}. LiDAR SLAM provides an alternative near quays and bridges, but depends on sufficient above-water structure~\cite{shen2023lidarusv}, which open water may not provide. For autonomous underwater vehicles (AUVs), GNSS is unavailable while submerged, and visual odometry can be degraded by turbidity, poor illumination and limited visibility~\cite{ferrera2019visual}. These limitations motivate navigation using onboard observations of the underwater environment.

Forward-looking sonar (FLS) provides acoustic imagery for inspection and mapping in conditions that challenge optical sensing~\cite{FERREIRA2015212}. Using these observations for odometry, however, requires accounting for their geometry. An FLS measures metric slant range and bearing but does not resolve elevation within its vertical beam. Consequently, interpreting image displacement as horizontal motion can introduce feature-dependent scale errors: the conversion depends on the unknown position of each return within the beam. Previous methods address this geometry through planar approximations, additional navigation sensors or multiview estimation~\cite{hurtos2015fourier,li2018posegraph,westman2020degeneracy,cafaro2026ifac}. Here, we investigate how the seabed return itself can constrain feature projection and how departures from the assumed seabed geometry affect the estimated motion.

We present \brina{} (Bottom-Referenced Inertial Navigation with AKAZE), a sonar--inertial SLAM system for marine vehicles using the FLS as its only exteroceptive sensor. Under a locally horizontal seabed model, the range
of the first seabed return, termed the \emph{seabed edge}, provides an altitude estimate using the known beam geometry and IMU-estimated roll and pitch. This altitude is used to project static seabed features into
horizontal coordinates before motion estimation. The resulting odometry is fused with IMU preintegration and loop closures in a factor graph. The sonar observations also support 3-D seabed mapping. The approach
targets seabed-observing survey motion on surface and underwater platforms.

\subsection{Contributions}\label{sec:contrib}

The main contributions are:
\begin{enumerate}
    \item \brina{}, a sonar--inertial SLAM system that combines seabed-edge altitude estimation, ground-projected AKAZE odometry with adaptive keyframe selection, and loop closures using sonar mosaics and seabed submaps.

    \item An analysis of the elevation-induced scale error. We identify motion conditions under which inertial measurements cannot separate scale from accelerometer bias, and derive a model and bound relating the remaining odometry error to seabed tilt and slope variability.

    \item Extensive simulation and experimental results: evaluation with an AUV in simulation over ten seabeds spanning tilt and roughness, and with USVs in ten field missions at six sites. The field trials use a BlueView at two frequencies and an Oculus in calm water and waves. Together, these evaluations assess the error model, trajectory accuracy and loop closures.
\end{enumerate}

\subsection{Related work}\label{sec:related}

\textbf{Acoustic navigation.}
Sonar-based navigation encompasses several sensing geometries and sensor combinations. Imaging-sonar SLAM has fused acoustic constraints with odometry from a Doppler velocity log (DVL) and an IMU~\cite{li2018posegraph}, while camera--sonar methods use acoustic measurements to recover monocular scale~\cite{yang2022scale}. Side-scan sonar correspondences have constrained AUV trajectories through loop closures added to dead reckoning~\cite{zhang2024sidescan}. Multibeam echosounder bathymetry has also supported SLAM with an IMU and no DVL~\cite{gong2025bathymetric}. These approaches establish acoustic observations as navigation measurements, while differing in the geometry measured and the additional information required.

\textbf{FLS odometry and elevation.}
Image-registration methods support FLS mosaicking~\cite{hurtos2015fourier}, while multiview formulations estimate motion and scene structure while accounting for sonar degeneracies~\cite{westman2020degeneracy}. Franchi et al.~\cite{franchi2020fls} also demonstrate DVL-free navigation by estimating velocity from FLS imagery within an adaptive navigation filter. Our focus is the seabed reference used to correct feature projection before registration and the scale error that remains when the seabed departs from that reference.

\textbf{Seabed projection and the seabed edge.}
The closest geometric precedent is Henson and Zakharov~\cite{henson2019attitude}, who estimate attitude and trajectory using a planar scene model. Their extension estimates pitch and altitude from the first and last illuminated points, assuming a flat scene and known vertical beamwidth. \brina{} combines the seabed edge with IMU attitude, avoiding the need to detect the far boundary of the footprint. It integrates this altitude estimate into sonar--inertial SLAM and quantifies how seabed geometry affects odometric scale. Compared with previous sonar-plane odometry~\cite{cafaro2026ifac}, it replaces a fixed image-to-motion conversion with altitude-dependent ground projection.

\Cref{sec:sonar} introduces the measurement model and scale problem; \cref{sec:brina} presents the system and error analysis; \cref{sec:exp} reports the experiments; and \cref{sec:concl} concludes.
\section{Sonar measurement model}\label{sec:sonar}
\begin{figure}[t]
\centering
\includegraphics[width=\columnwidth]{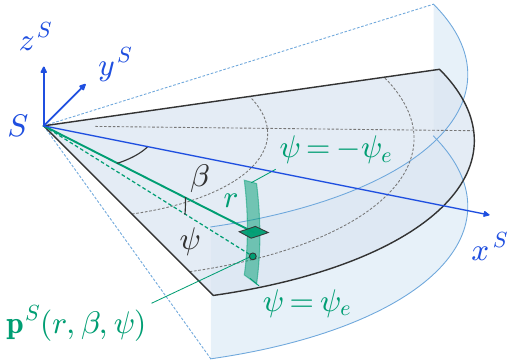}
\caption{Geometry of a forward-looking sonar in its frame $S$ ($x^S$ along the boresight, $y^S$ to the
left, $z^S$ up). The fan outlined in black is the image plane, $\psi=0$, and the light blue volume is the
vertical beam, between its lower edge $\psi=\psi_e$ and its upper edge $\psi=-\psi_e$. A pixel (green
square) stores only the slant range $r$ and the bearing $\beta$, so its echo may come from any point of
the green arc: the dot is the actual return $\vect{p}^{S}(r,\beta,\psi)$, at an
elevation $\psi$ that the sonar does not measure.}
\label{fig:geometry}
\end{figure}
In this section, we introduce the geometry and measurement model of an FLS. We first show that an FLS measurement does not uniquely determine the 3-D position of a return because its elevation angle is unobserved. We then show that, under assumptions on the seabed and on the observed features, the sonar altitude provides the additional geometric constraint required to recover their elevation and project them onto the seabed. This projection will later be used to explain the scale error that arises when motion is estimated directly in the sonar image plane.

\subsection{The sonar image}\label{sec:sonar-image}

Let $r$ be the slant range of a return, $\beta$ its bearing in the fan and $\psi$ its elevation inside
the vertical beam, positive upwards from the centre of the fan. In the sonar frame $S$ the return lies at
\begin{equation}
\vect{p}^{S}(r,\beta,\psi)=r\,[\cos\psi\cos\beta,\ \cos\psi\sin\beta,\ \sin\psi]^\top ,
\label{eq:fls}
\end{equation}
the spherical model of an imaging sonar \cite{hurtos2015fourier,westman2020degeneracy}, with the $x$ axis
of $S$ along the boresight. The sonar image, however, stores only $r$ and $\beta$; the elevation $\psi$
within the vertical beam is not measured. Therefore, a pixel does not correspond to a unique point in
3-D space, but to an arc of possible positions parameterised by $\psi$ (\cref{fig:geometry}). This is
the elevation ambiguity of an imaging sonar.

In our application, the sonar is mounted on the vehicle, pitched down by the \emph{tilt} $\alpha$ and offset from the body frame $B$,
the IMU frame (forward, left, up), by the lever arm $\vect{t}_{BS}$. The position of the same return in
the body frame is therefore
\begin{equation}
\vect{p}^{B}=\mat{R}_y(\alpha)\,\vect{p}^{S}+\vect{t}_{BS} ,
\label{eq:body}
\end{equation}
with $\mat{R}_y(\alpha)$ the rotation of the mount about the lateral axis.

By itself, \eqref{eq:body} does not remove the elevation ambiguity, since $\psi$ remains unknown. To
recover the elevation of the features used for odometry, we make two assumptions about the observed
scene:
\begin{itemize}
    \item The seabed within the footprint of a ping is locally flat and horizontal.
    \item The features detected and tracked in the sonar image correspond to static points lying on that seabed.
\end{itemize} 
Let the \emph{sonar altitude} $h$ be the height of the sonar above the local seabed plane. Under these
assumptions, a feature measured at $(r,\beta)$ must lie both on the arc of possible positions defined by
the sonar measurement and on the seabed plane. The known altitude therefore supplies the missing
constraint needed to determine which point on that arc corresponds to the observed feature. Taking the
vertical component of \eqref{eq:body} and imposing the seabed-plane constraint gives
\begin{equation}
r\,(\sin\psi\cos\alpha-\cos\psi\cos\beta\sin\alpha)=-h ,
\label{eq:flat}
\end{equation}
which relates the unobserved elevation $\psi$ to the measured range and bearing and to the known sonar
altitude.

If $h$ is known, \eqref{eq:flat} can therefore be solved for the elevation of each seabed feature. The
solution inside the vertical fan is
\begin{equation}
\begin{aligned}
\psi^{\star}(r,\beta;h)=\;&\operatorname{atan2}(\cos\beta\sin\alpha,\cos\alpha)\\
&-\arcsin\frac{h}{r\sqrt{\cos^2\alpha+\cos^2\beta\sin^2\alpha}} .
\end{aligned}
\label{eq:psistar}
\end{equation}
Thus, although elevation is not measured directly by the sonar, the combination of the sonar
measurement $(r,\beta)$, the known altitude $h$, and the assumption that the feature lies on the
locally horizontal seabed determines its elevation $\psi^\star$.

Substituting \eqref{eq:psistar} in \eqref{eq:fls} and \eqref{eq:body} gives the \emph{ground
projection} $\vect{\pi}$, the horizontal position of the feature in the body frame,
\begin{equation}
\vect{\pi}(r,\beta;h)\!=\!\begin{bmatrix} r(\cos\psi^{\star}\cos\beta\cos\alpha+\sin\psi^{\star}\sin\alpha)\\ r\cos\psi^{\star}\sin\beta\end{bmatrix}\!+\!\begin{bmatrix}t_{BS,x}\\ t_{BS,y}\end{bmatrix} .
\label{eq:ground}
\end{equation}
The ground projection therefore converts a sonar measurement, which is naturally expressed in slant
range and bearing, into a metric position on the local seabed plane. Since the projected features are
assumed static, differences between their projected positions in consecutive pings can then be used to
estimate the motion of the vehicle.

This relationship is particularly clear for the centre beam, $\beta=0$. In this case,
the plane condition \eqref{eq:flat} and the first component of \eqref{eq:ground} without the lever arm
become
\begin{equation}
h=r\sin(\alpha-\psi^\star),\qquad g=r\cos(\alpha-\psi^\star) ,
\label{eq:centre}
\end{equation}
where $g$ is the \emph{ground distance} of the seabed point, the horizontal distance from the sonar to the
point. The altitude $h$, ground distance $g$, and slant range $r$ form a right triangle. Consequently,
\begin{equation}
g=\sqrt{r^2-h^2} .
\label{eq:X}
\end{equation}
Equation~\eqref{eq:X} shows explicitly that slant range is not the same as horizontal distance. The
conversion between the two depends on the sonar altitude. Therefore, a displacement measured as a
change of range in the sonar image cannot in general be interpreted directly as the metric displacement
of the vehicle along the seabed.

For a vehicle moving forward over a flat seabed, $h$ remains constant while $g$ changes with the
relative displacement between the sonar and the observed static seabed point. Differentiating
$r^2=g^2+h^2$ at constant $h$ gives
\begin{equation}
\frac{\partial r}{\partial g}=\frac{g}{r}=\sqrt{1-\frac{h^2}{r^2}} ,
\label{eq:texture}
\end{equation}
which shows that the same metric displacement on the seabed produces different changes in image range
depending on the range of the observed feature. In particular, the image texture moves less at short
range than at long range. The ground projection removes this range-dependent scaling by mapping the
returns onto the seabed before their displacement is estimated.

The remaining quantity required for this projection is the sonar altitude $h$. In the configuration
considered here, it can be estimated directly from the FLS measurement itself. Along every beam the
seabed first appears at the \emph{seabed edge}, the range $r_e$ at which the echo rises from
the dark water column to the seabed level. Under the locally flat seabed
assumption, this first echo is generated by the steepest downward-looking ray of the vertical beam,
since it is the first ray to intersect the seabed. Its elevation, $\psi_e$, corresponds to the lower
edge of the vertical beam (\cref{fig:geometry}) and is therefore known from the transducer geometry.

Reading \eqref{eq:flat} at $\psi=\psi_e$, in the centre beam, gives the altitude from the
range of the edge,
\begin{equation}
h=r_e\sin(\alpha-\psi_e) ,
\label{eq:altitude}
\end{equation}
the depth reached by a ray of length $r_e$ that points $\alpha-\psi_e$ below the horizontal. The point
where that ray meets the seabed is the nearest point of the footprint, at the ground distance of the
edge
\begin{equation}
g_e=h\cot(\alpha-\psi_e) .
\label{eq:Xe}
\end{equation}
Thus, the seabed edge provides the altitude required by \eqref{eq:psistar}--\eqref{eq:ground} directly
from the sonar image. The complete geometric argument is therefore as follows: the FLS measures range
and bearing but not elevation; assuming that a detected feature lies on a locally horizontal seabed
allows its elevation to be recovered when the sonar altitude is known; assuming that the feature is
static then allows the displacement of its ground projection between consecutive pings to be attributed
to the motion of the vehicle.

\subsection{Motion in the sonar plane}\label{sec:sonar-scale}

We now consider what happens when vehicle motion is estimated directly from the sonar image without
resolving the elevation ambiguity described in \cref{sec:sonar-image}. In this case, the detected
features are treated as if they lay in the sonar plane, which is equivalent to assuming zero elevation,
$\psi=0$. This assumption allows image displacements to be converted into metric motion using the known
mounting geometry, but it is generally incorrect for seabed features, whose actual elevation within the
vertical beam depends on their position relative to the sonar and on the sonar altitude. As a result,
motion estimated in the sonar plane carries an unknown scale.

The \emph{step} is the motion of the vehicle between two pings, usually consecutive, a planar translation
and a rotation, and the \emph{step length} $d$ is the distance travelled between them. Without knowing the
altitude of the sonar over the seabed, the elevation of a matched feature cannot be recovered and the
step can only be inferred from its displacement in the sonar image. To expose the resulting scale error,
consider a seabed point in the centre beam at ground distance $g$ and slant range
$r=\sqrt{g^2+h^2}$ from \eqref{eq:X}, and let the step be a pure translation of length $d$ towards it.
The altitude $h$ does not change, so the ground distance of the point decreases from $g$ to $g-d$ and
its slant range from $r(g)$ to $r(g-d)$.

The step length is small compared with the ground distance, so the decrease of the slant range is, to
first order in $d$, the derivative \eqref{eq:texture} times the step length,
\begin{equation}
\begin{aligned}
\Delta r&=r(g)-r(g-d)=\sqrt{g^2+h^2}-\sqrt{(g-d)^2+h^2}\\
&\approx\frac{\partial r}{\partial g}\,d=\frac{g}{r}\,d=\cos(\alpha-\psi)\,d ,
\end{aligned}
\label{eq:dr}
\end{equation}
where the last equality is \eqref{eq:centre}: $g/r$ is the cosine of the depression of the point below
the horizontal, $\alpha-\psi$, the tilt of the mount minus the elevation of the point inside the vertical
beam.

Equation~\eqref{eq:dr} shows the source of the ambiguity. The range displacement observed in the sonar
image is not equal to the vehicle displacement $d$; it is the projection of that displacement according
to the depression angle of the observed feature. Recovering $d$ from $\Delta r$ therefore requires the
feature elevation $\psi$, which the imaging sonar does not measure. Different seabed features can lie at
different elevations inside the vertical beam and therefore produce different range changes for the same
vehicle displacement.

A sonar-plane front end avoids this unknown by imposing $\psi=0$. In other words, it assumes that the
matched feature lies on the centre plane of the vertical beam, so that its depression angle is taken to
be the known sonar tilt $\alpha$. A front end that works in the sonar plane, such as the
odometry in~\cite{cafaro2026ifac}, therefore reports the step length
\begin{equation}
\hat d=\frac{\Delta r}{\cos\alpha}=s\,d,\qquad s=\frac{\cos(\alpha-\psi)}{\cos\alpha}=\frac{1}{\cos\alpha}\sqrt{1-\frac{h^2}{r^2}} ,
\label{eq:scale-slant}
\end{equation}
where the last form follows from \eqref{eq:dr} and \eqref{eq:texture}. The estimated displacement
$\hat d$ is thus related to the true displacement $d$ by the scale factor $s$. The scale is one only
when the zero-elevation assumption is correct, i.e., for a point on the boresight, $\psi=0$. For every
other point, $s$ depends on the actual elevation of the matched feature and therefore on its geometry
relative to the seabed and the sonar altitude. Since this elevation is not available to a sonar-plane
front end, the metric scale of the estimated motion is unknown. Off the centre beam $s$ also depends on
the bearing; the centre-beam case is sufficient to expose the ambiguity.

This also shows why the error cannot, in general, be represented by a fixed calibration factor. The
elevation of a seabed feature changes with its range and with the geometry of the observed bottom.
Consequently, different matched features, and even the same feature observed from different vehicle
positions, can induce different values of $s$. The scale error is therefore a consequence of neglecting
the three-dimensional geometry of the sonar measurement rather than of a constant sensor calibration
error.

The IMU measures in metres per second squared, so in principle it can supply the missing scale, as it
does in visual--inertial odometry~\cite{forster2017onmanifold}. However, inertial measurements can
separate scale from accelerometer bias only when the vehicle motion provides sufficient excitation.
Proposition~1 states this for the planar motion of a surface vessel, following the observability analyses
of visual--inertial odometry~\cite{martinelli2012closed,wu2017vinswheels}.

\textbf{Proposition 1} (after~\cite{martinelli2012closed,wu2017vinswheels})\textbf{.} Consider planar
motion without sideways slip, with forward speed $v(t)$ and yaw rate $\omega(t)$ known from the gyro,
the accelerometer at the centre of rotation with gravity removed through the attitude, accelerometer
biases $b_x,b_y$ constant over a window of time, and a sonar front end that measures the speed up to an
unknown scale, $\hat v(t)=s\,v(t)$. The accelerometer measures
$a_x=\dot v+b_x$ and $a_y=v\,\omega+b_y$. Then $s$ cannot be identified from $\hat v$, $\omega$, $a_x$
and $a_y$ on the window if and only if $\dot{\hat v}(t)$ and $\hat v(t)\,\omega(t)$ are both constant
on it.

\emph{Proof.} Writing the true speed as $v=\hat v/s$, the accelerometer measures
\begin{equation}
a_x=\frac{\dot{\hat v}}{s}+b_x,\qquad a_y=\frac{\hat v\,\omega}{s}+b_y .
\label{eq:p1-meas}
\end{equation}
Suppose a second set $(s',b_x',b_y')$ with $s'\neq s$ explains the same data. Subtracting its two
equations from \eqref{eq:p1-meas}, for all $t$ in the window,
\begin{equation}
\Big(\frac1s-\frac1{s'}\Big)\dot{\hat v}(t)=b_x'-b_x,\qquad \Big(\frac1s-\frac1{s'}\Big)\hat v(t)\,\omega(t)=b_y'-b_y .
\label{eq:p1-diff}
\end{equation}
The right-hand sides of \eqref{eq:p1-diff} are constant and $1/s-1/s'\neq0$, hence
\begin{equation}
\dot{\hat v}(t)=\text{const},\qquad \hat v(t)\,\omega(t)=\text{const} .
\label{eq:p1-const}
\end{equation}
Conversely, if \eqref{eq:p1-const} holds, every $s'$ explains the data \eqref{eq:p1-meas} with the biases
\begin{equation}
b_x'=b_x+\Big(\frac1s-\frac1{s'}\Big)\dot{\hat v},\qquad b_y'=b_y+\Big(\frac1s-\frac1{s'}\Big)\hat v\,\omega .\qquad\square
\label{eq:p1-bias}
\end{equation}

Proposition~1 shows that, under insufficiently excited motion, an error in the scale of the sonar
odometry can be absorbed by a different estimate of the accelerometer biases. The IMU therefore cannot
in general correct the scale error introduced by the zero-elevation assumption during the smooth,
predominantly planar motions that constitute a USV survey. The missing scale must instead be recovered
from the sonar geometry itself. Once the sonar altitude over the seabed is known, the elevation of the
static seabed features can be estimated using \eqref{eq:psistar}, and the features can be projected onto
the seabed using \eqref{eq:ground}. Motion can then be estimated from these ground-projected features
rather than directly in the elevation-ambiguous sonar plane. How \brina{} estimates the altitude ping by
ping and uses this projection in the odometry is described in \cref{sec:brina}.
\section{\brina{}}\label{sec:brina}
This section presents \brina{}, the proposed sonar--inertial SLAM system for marine vehicles. \brina{} uses the FLS as its only exteroceptive sensor and exploits the seabed as a metric reference for both odometry and mapping. The method combines inertial propagation, bottom-referenced sonar odometry, and loop closures within a single factor graph. The estimation framework is implemented in GTSAM and follows the factor-graph formulation of~\cite{dellaert2017factor,gtsam43}, using on-manifold IMU preintegration~\cite{forster2017onmanifold}. Loop-closure constraints are made robust to inconsistent detections using pairwise consistency maximisation~\cite{mangelson2018pcm} and robust kernels, following their use for robust loop-closure handling in multi-robot SLAM \cite{lajoie2020doorslam,tian2022kimeramulti}.

The remainder of this section is organised as follows. \Cref{sec:overview} gives an overview of the complete system and the interaction between its components. \Cref{sec:f0} introduces the IMU factor F0. \Cref{sec:f1} describes the sonar odometry factor F1, including the estimation of the sonar altitude from the seabed edge, while \cref{sec:f1-error} analyses the odometry error introduced when the estimated altitude does not exactly represent the geometry of the observed seabed. \Cref{sec:f2} presents the loop-closure factor F2, and \cref{sec:map} describes the factor graph that combines the three sources of information into the final SLAM estimate.

\subsection{Overview}\label{sec:overview}

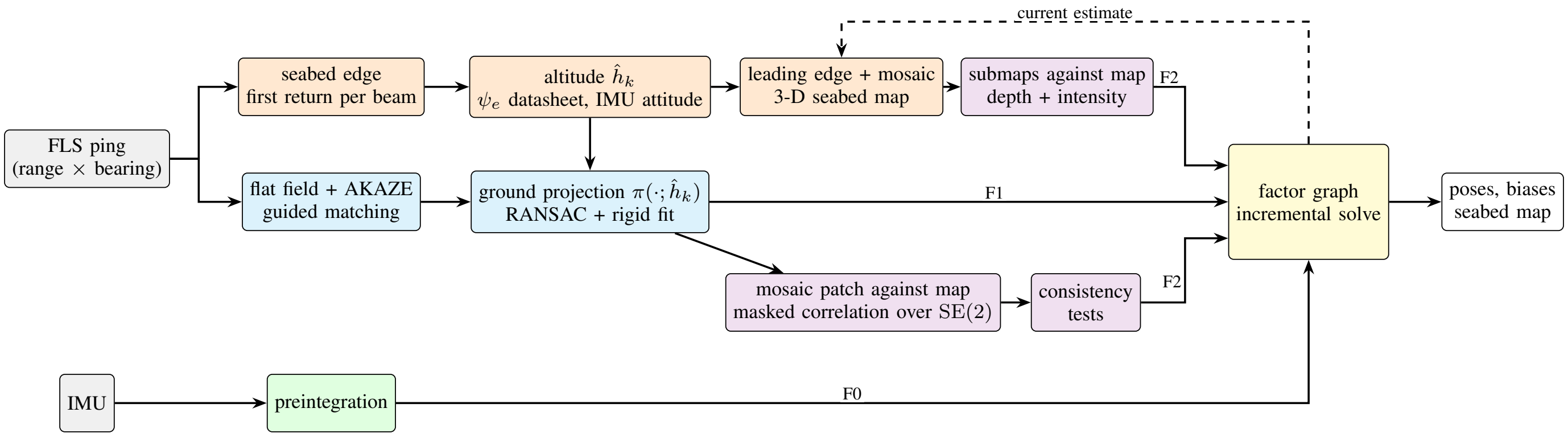
\begin{figure*}[t]
\centering
\resizebox{\textwidth}{!}{%
\begin{tikzpicture}[x=1cm,y=1cm,
  box/.style={draw, rounded corners=2pt, align=center, font=\footnotesize, minimum height=8mm, inner sep=3pt, fill=white},
  inp/.style={box, fill=gray!12}, le/.style={box, fill=orange!18}, ak/.style={box, fill=cyan!12},
  imu/.style={box, fill=green!12}, lc/.style={box, fill=violet!12}, fg/.style={box, fill=yellow!20, minimum height=16mm},
  arr/.style={-{Stealth[length=2mm]}, thick}, lab/.style={font=\scriptsize, fill=white, inner sep=1pt}]
\node[inp] (fls) at (0,0.6) {FLS ping\\(range $\times$ bearing)};
\node[le] (le) at (3.4,1.6) {seabed edge\\first return per beam};
\node[le] (alt) at (7.0,1.6) {altitude $\hat h_k$\\$\psi_e$ datasheet, IMU attitude};
\node[le] (map) at (10.5,1.6) {leading edge + mosaic\\3-D seabed map};
\node[lc] (sub) at (13.5,1.6) {submaps against map\\depth + intensity};
\node[ak] (ak) at (3.4,0.0) {flat field + AKAZE\\guided matching};
\node[ak] (gp) at (7.0,0.0) {ground projection $\vect{\pi}(\cdot;\hat h_k)$\\RANSAC + rigid fit};
\node[lc] (patch) at (10.8,-1.4) {mosaic patch against map\\masked correlation over $\SE{2}$};
\node[lc] (pcm) at (13.9,-1.4) {consistency\\tests};
\node[inp] (imu) at (0,-2.8) {IMU};
\node[imu] (pre) at (3.4,-2.8) {preintegration};
\node[fg] (graph) at (17.0,0.0) {factor graph\\incremental solve};
\node[box] (out) at (19.7,0.0) {poses, biases\\seabed map};
\draw[arr] (fls.east) -- ++(0.4,0) |- (le.west); \draw[arr] (fls.east) -- ++(0.4,0) |- (ak.west);
\draw[arr] (le) -- (alt); \draw[arr] (ak) -- (gp); \draw[arr] (alt) -- (gp); \draw[arr] (alt) -- (map); \draw[arr] (gp) -- (patch);
\draw[arr] (gp.east) -- node[lab, above, pos=0.55]{F1} (graph.west);
\draw[arr] (map) -- (sub);
\draw[arr] (sub.east) -| node[lab, above, pos=0.25]{F2} ([xshift=-6mm]graph.north west) -- ([xshift=-6mm, yshift=-3mm]graph.north west) -- ([yshift=-3mm]graph.north west);
\draw[arr] (patch) -- (pcm);
\draw[arr] (pcm.east) -| node[lab, left, pos=0.75]{F2} ([xshift=-6mm]graph.south west) -- ([xshift=-6mm, yshift=3mm]graph.south west) -- ([yshift=3mm]graph.south west);
\draw[arr] (imu) -- (pre); \draw[arr] (pre.east) -| node[lab, above, pos=0.25]{F0} (graph.south); \draw[arr] (graph) -- (out);
\draw[arr, dashed] (graph.north) |- node[lab, above, pos=0.75]{current estimate} ([yshift=5mm]map.north) -- (map.north);
\end{tikzpicture}}
\caption{\brina{} pipeline. The grey boxes are the two inputs, the sonar ping and the IMU. The orange
boxes are the seabed-edge chain: the first return of each beam gives the altitude of the ping and,
through the leading edge, the 3-D seabed map. The cyan boxes are the feature chain: AKAZE features
projected onto the seabed at that altitude give the step between pings, factor F1. The violet boxes are
the two loop closures on the map, factor F2: submaps of depth and intensity (top) and a patch of the
mosaic (bottom). The green box is the IMU preintegration, factor F0. The yellow
box is the factor graph, and the dashed arrow the current estimate that places the leading edge in the
map.}
\label{fig:pipeline}
\end{figure*}

\Cref{fig:pipeline} summarises the \brina{} pipeline. A state is associated with every sonar ping $k$ and contains the body pose $\mat{X}_k\in\SE{3}$, the velocity $\vect{v}_k$, and the IMU biases $\vect{b}_k$. These states are constrained by three complementary factors.

F0 is the IMU preintegration factor between consecutive sonar pings. It propagates the vehicle dynamics between states, constrains the relative rotation and velocity, and allows the IMU biases to be estimated jointly with the trajectory.

F1 provides the bottom-referenced sonar odometry. For each ping, the seabed edge is used to estimate the sonar altitude, and AKAZE features detected beyond it are projected onto the seabed using this altitude. The pings used for motion estimation are selected dynamically as keyframes according to the amount of usable information available for registration, rather than at a fixed temporal spacing. Features matched between successive keyframes are registered through a planar rigid transformation to estimate their relative motion. In regions with rich and persistent seabed texture, keyframes can be selected close in time; when the available information is insufficient, the keyframe interval is adapted until a reliable relative-motion estimate can be formed. Because the registration is performed after projecting the matched features onto the seabed, F1 provides the metric translational constraint that resolves the scale ambiguity discussed in \cref{sec:sonar-scale}.

F2 provides loop-closure constraints when the vehicle revisits previously observed areas. Two complementary loop-closure mechanisms are used:
\begin{itemize}
\item \textbf{Sonar-mosaic loop closure:} patches of the current sonar mosaic are matched against previously observed mosaic regions to detect revisits from the acoustic intensity pattern.
\item \textbf{Seabed-submap loop closure:} local submaps containing seabed depth and acoustic intensity are registered against earlier portions of the map within the region allowed by the accumulated odometry uncertainty.
\end{itemize}
The two mechanisms exploit complementary representations of the same environment and provide constraints that correct the drift accumulated by F0 and F1 over longer trajectories.
The three factors are combined in a single factor graph and optimised incrementally as new measurements arrive. When a loop closure is detected, the graph is re-optimised so that the correction is propagated through the trajectory and the seabed map.

\subsection{IMU factor (F0)}\label{sec:f0}
The IMU factor F0 constrains the vehicle motion between two consecutive sonar pings. Let $i$ and $k$
denote two consecutive ping times. The IMU measurements collected between them are preintegrated on the
manifold~\cite{forster2017onmanifold} into the relative rotation, velocity, and position increments
$(\Delta\meas{\mat{R}}_{ik},\Delta\meas{\vect{v}}_{ik},\Delta\meas{\vect{p}}_{ik})$. Preintegration
summarises all IMU samples between the two states into a single relative-motion constraint while
accounting for the current estimate of the IMU biases.

The resulting factor compares the preintegrated measurements with the motion predicted by the states at
pings $i$ and $k$,
\begin{equation}
\vect{e}_{\mathrm{F0}}=\vect{r}_{\mathrm{IMU}}\big(\mat{X}_i,\vect{v}_i,\vect{b}_i,\mat{X}_k,\vect{v}_k;\ \Delta\meas{\mat{R}}_{ik},\Delta\meas{\vect{v}}_{ik},\Delta\meas{\vect{p}}_{ik}\big),
\label{eq:f0}
\end{equation}
where $\vect{r}_{\mathrm{IMU}}$ is the preintegrated residual of~\cite{forster2017onmanifold}. The
residual accounts explicitly for gravity and for the dependence of the preintegrated measurements on
the IMU biases. The biases are included in the graph state and modelled as slowly varying quantities,
allowing them to be estimated jointly with the vehicle trajectory.

In addition to the inertial constraint, the motion of a surface vehicle provides a simple geometric constraint on
the vertical component of the trajectory. Since the vehicle remains on the water surface, its heave is
assumed negligible over the missions considered in this work. The vertical position of each pose is
therefore constrained around zero through the prior
\begin{equation}
e_{\mathrm{S},k}=\frac{p_{k,z}}{\sigma_{\mathrm{S}}} ,
\label{eq:heave}
\end{equation}
with $p_{k,z}$ the vertical component of the position and $\sigma_{\mathrm{S}}$ its standard deviation.
The parameter $\sigma_{\mathrm{S}}$ controls the strength of the surface constraint and allows small
vertical deviations from the nominal zero-heave condition. For an underwater vehicle, the vertical position is referred to the seabed instead: $p_{k,z}$ in \eqref{eq:heave} is replaced by its difference from the sonar altitude estimated in \cref{sec:f1}.

\subsection{Sonar factor (F1)}\label{sec:f1}

F1 provides the bottom-referenced relative motion between two dynamically selected sonar pings $i$ and $k$, with $i<k$. We refer to these pings as the two keyframes of an F1 \emph{step}: $i$ is the reference keyframe and $k$ is the next keyframe selected from the subsequent pings. Every sonar ping still has a state in the graph, but an F1 constraint is introduced only when a sufficiently informative pair $(i,k)$ has been found.

The measurement produced by F1 is the planar rigid motion
$\meas{\vect{z}}_{ik}=(\vect{t}_{ik},\theta_{ik})$, where
$\vect{t}_{ik}=(t_x,t_y)$ is the translation expressed in the body frame of ping $i$ and
$\theta_{ik}$ is the rotation about the vertical axis. The construction of this measurement proceeds
through five stages. First, the seabed edge provides an estimate of the sonar altitude at each ping.
Second, this altitude is used to project the detected sonar features onto the seabed. Third, features
between the reference ping $i$ and a candidate ping $k$ are matched. Fourth, the matched seabed points
are fitted with a planar rigid transformation. Finally, the quality of this fit determines whether the
candidate ping becomes the next keyframe $k$.

\begin{figure}[t]
\centering
\includegraphics[width=\columnwidth]{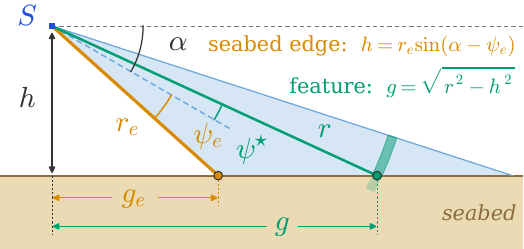}
\caption{How F1 becomes metric, in the vertical plane of the centre beam ($\beta=0$). The seabed edge
(orange) is the echo of the lower edge of the vertical beam, at the elevation $\psi_e$ from the centre of
the fan (dashed): its range $r_e$ gives the altitude $h$ of the sonar $S$ and the ground distance $g_e$ where
the footprint starts. A feature (green) at the slant range $r$ lies somewhere on its arc across the beam; the
seabed selects its elevation $\psi^\star$ and its ground distance $g$.}
\label{fig:edge}
\end{figure}

\textbf{Altitude.}
The first quantity required to project the sonar measurements onto the seabed is the altitude of the
sonar. In every beam $\beta$, the seabed edge is the range $r_e(\beta)$ at which the echo rises from
the water-column level to the seabed level and remains above it; beams without such a rise provide no
edge measurement. Each valid beam gives an independent altitude estimate from the depth reached by its
edge ray,
\begin{equation}
h_k(\beta)=r_e(\beta)\,u_z(\beta) ,
\label{eq:hbeam}
\end{equation}
where $u_z(\beta)$ is the downward component of the unit edge ray $\vect{u}(\beta,\psi_e)$ of
\eqref{eq:fls} rotated by the tilt and by the roll and pitch of the ping. In the centre beam of a level
vehicle $u_z=\sin(\alpha-\psi_e)$ and \eqref{eq:hbeam} reduces to \eqref{eq:altitude}, the case drawn in
\cref{fig:edge}.

Since an individual seabed-edge measurement can be missing or corrupted, the beam estimates are first
combined within the ping. The raw altitude of ping $k$ is taken as the median over the central beams,
\begin{equation}
\tilde h_k=\operatorname{med}_{|\beta|<10^\circ}h_k(\beta) ,
\label{eq:hraw}
\end{equation}
which rejects isolated beams with a spurious or missing edge. The resulting ping altitude is then
smoothed temporally by taking the median over a window of $w$ pings on either side,
\begin{equation}
\hat h_k=\operatorname{med}\{\tilde h_q:\ |q-k|\le w\} .
\label{eq:Hk}
\end{equation}
The window spans only a few seconds, which is short compared with the time scale over which the altitude
of a surface vessel changes. It therefore suppresses ping-to-ping noise and isolated erroneous
detections while retaining the slowly varying altitude. The same procedure provides the altitude for
the reference ping $i$.

\textbf{Seabed points.}
Once the altitude is available, the sonar features can be moved from the elevation-ambiguous image
geometry to the seabed plane. Before feature extraction, the stationary pattern produced by the sonar
head is removed. Let $I_k$ be the recorded intensity image and
$L_k=\log(1+I_k)$ its logarithmic representation. The pattern that remains fixed in the sonar frame is
estimated from the surrounding pings $\mathcal{W}_k$ and subtracted,
\begin{equation}
F_k=L_k-\frac{1}{|\mathcal{W}_k|}\sum_{q\in\mathcal{W}_k}L_q .
\label{eq:flatfield}
\end{equation}

AKAZE~\cite{alcantarilla2013akaze} features and binary descriptors are then extracted from the
contrast-equalised image $F_k$. Features are detected only beyond the seabed edge of each beam, where
the sonar return corresponds to the seabed. When more features are available than the matching stage
requires, only those with the strongest detector responses are retained.

A feature $m$ detected at sub-pixel range bin $\ell_m$ and beam $c_m$ has slant range
$r_m=r(\ell_m)$ and bearing $\beta_m=\beta(c_m)$. The altitude estimate \eqref{eq:Hk} now supplies the
quantity that was missing in the sonar measurement model. Using the ground projection
\eqref{eq:ground}, together with the roll $\phi_k$ and pitch $\vartheta_k$ of the vehicle, the feature is
placed on the seabed in the body frame of ping $k$,
\begin{equation}
\vect{b}_m=\vect{\pi}(r_m,\beta_m;\hat h_k,\phi_k,\vartheta_k)\in\real^2 .
\label{eq:seabedpt}
\end{equation}
\Cref{fig:edge} shows this projection in the centre beam, where it reduces to the ground distance
\eqref{eq:X}. The features of the reference ping $i$ are projected in the same way, giving the corresponding points
$\vect{a}_m$. Under the static-seabed assumption introduced in \cref{sec:sonar-image}, a feature observed
in both pings corresponds to the same physical point. Its two ground-projected coordinates are therefore
related only by the relative planar motion between the two body frames,
\begin{equation}
\vect{a}_m=\mat{R}(\theta_{ik})\,\vect{b}_m+\vect{t}_{ik} ,
\label{eq:rigid}
\end{equation}
and, because $\vect{a}_m$ and $\vect{b}_m$ are expressed in metres on the seabed, the translation
$\vect{t}_{ik}$ is also metric.

\textbf{Matching.}
Equation~\eqref{eq:rigid} defines the geometric relation that a correct feature correspondence must
satisfy. The next step is therefore to identify pairs of projected features that are likely to represent
the same seabed point.

Sonar imagery contains substantial speckle, and many features may have similar binary descriptors.
\brina{} consequently restricts the descriptor search using a prediction of the relative motion between
the two pings. The rotation is predicted by integrating the gyro measurements between $i$ and $k$,
while the translational prediction assumes that the vehicle approximately maintains its speed while
following the measured turn. This prediction is used only to define the search region; it is not used
as the F1 motion measurement.

Feature $m$ of ping $i$ is paired with the feature $m'$ of ping $k$ whose predicted position lies
within a gate of $\vect{a}_m$ and whose descriptor $\vect{f}_{m'}$ has the smallest Hamming distance
from $\vect{f}_m$,
\begin{equation}
m'(m)=\operatorname*{arg\,min}_{m':\ \|\vect{a}_m-\hat{\vect{a}}_{m'}\|\le\varepsilon_{\mathrm{g}}}\ d_H(\vect{f}_m,\vect{f}_{m'}) ,
\label{eq:match}
\end{equation}
where $\hat{\vect{a}}_{m'}$ is the predicted position and $\varepsilon_{\mathrm{g}}$ is the gating radius, which grows
with the predicted step length. A correspondence is retained only when the best descriptor distance is
sufficiently separated from the second-best candidate.

The descriptor test alone does not guarantee geometrically consistent correspondences. RANSAC~\cite{fischler1981ransac}
is therefore applied to the candidate pairs, using
\eqref{eq:rigid} as the motion model. The inliers are the correspondences that agree with a common
planar rigid transformation within the tolerance $\varepsilon$.

\textbf{Fit.}
Given the $n$ RANSAC inliers, the relative motion between the two pings is obtained by fitting the rigid
model \eqref{eq:rigid}. The translation and rotation are the least-squares solution
\begin{equation}
\min_{\vect{t}_{ik},\theta_{ik}}\ \sum_{m=1}^{n}\big\|\vect{a}_m-\mat{R}(\theta_{ik})\vect{b}_m-\vect{t}_{ik}\big\|^2 ,
\label{eq:fit}
\end{equation}
which has a closed form~\cite{arun1987lsq}.

The geometry of an FLS makes this fit intrinsically anisotropic. Most detected features occupy a band
ahead of the sonar at a ground distance $g$. A small rotation $\epsilon_\theta$ moves these points
laterally by approximately $g\,\epsilon_\theta$, which can resemble a sideways translation. The fitted
yaw and lateral translation are therefore strongly correlated. A fit is retained only when a minimum
fraction of the candidate correspondences agrees with it and when its estimated rotation remains within
a few degrees of the gyro prediction.

The resulting relative-motion measurement $\meas{\vect{z}}_{ik}$ is assigned the least-squares
covariance~\cite{barfoot2017state}
\begin{equation}
\mat{\Sigma}_{z}=\kappa\,\hat{\sigma}^2\,(\mat{J}^\top\mat{J})^{-1}+\mat{\Sigma}_{\min} ,
\label{eq:akazecov}
\end{equation}
where $\hat\sigma$ is the residual standard deviation of the inlier pairs, $\mat{J}$ is the Jacobian of
the residuals of \eqref{eq:fit} with respect to $(t_x,t_y,\theta_{ik})$, $\kappa$ is an inflation factor
set once using a $\chi^2$ consistency test of the whitened residuals, and
$\mat{\Sigma}_{\min}$ is a covariance floor. The covariance therefore retains the correlation between
the components of the fitted motion rather than treating them independently.

\textbf{Step selection.}
The previous stages provide a candidate motion measurement for any pair $(i,k')$. The remaining
question is which subsequent ping $k'$ should become the endpoint $k$ of the F1 step.

Over a textured seabed, consecutive pings contain repeatable features and the fit between them already
provides a reliable motion estimate. Over fine speckle, however, the apparent image pattern can change
substantially between consecutive pings. Many matches produced by \eqref{eq:match} can then be
incorrect, and some of these wrong matches will agree with a rigid transformation simply by chance.
Since the search is confined to a gate of radius $\varepsilon_{\mathrm{g}}$, the expected number of wrong
correspondences falling within the inlier tolerance $\varepsilon$ is
\begin{equation}
\lambda_{ik}=|\mathcal{C}_{ik}|\,\Big(\frac{\varepsilon}{\varepsilon_{\mathrm{g}}}\Big)^{2} .
\label{eq:chance}
\end{equation}
Here $\mathcal{C}_{ik}$ is the set of candidate correspondences between pings $i$ and $k$.

This chance agreement is important because the matching gate is centred around the predicted motion.
A fit supported mainly by accidental correspondences therefore tends to reproduce the prediction rather
than measure the displacement of the seabed texture. The number of RANSAC inliers $n$ is consequently
compared with the expected chance count \eqref{eq:chance}. Treating this count as approximately Poisson,
its excess in standard deviations is
\begin{equation}
\xi_{ik}=\frac{n-\lambda_{ik}}{\sqrt{\lambda_{ik}}} ,
\label{eq:signif}
\end{equation}
and a fit is considered \emph{clear} when $\xi_{ik}\ge\xi_{\min}$.

The significance \eqref{eq:signif} is also the quantity used to choose the next F1 keyframe. Starting
from the reference ping $i$, each following ping is considered in temporal order and processed through
the matching and fitting stages above. The step terminates at the first ping for which the fitted motion
is sufficiently significant,
\begin{equation}
k=\min\big\{k'>i:\ \Delta t_{ik'}\ge\Delta t_{\min},\ \xi_{ik'}\ge\xi_{\min}\big\} ,
\label{eq:step}
\end{equation}
where $\Delta t_{ik'}$ is the time elapsed between the two pings and $\Delta t_{\min}$ is the minimum
allowed step duration.

This rule makes the F1 keyframe spacing adaptive to the information contained in the sonar image. Over
a well-textured seabed, the next ping already provides a clear fit and becomes $k$. Over speckle, the
reference ping $i$ is retained while subsequent pings are tested until the accumulated vehicle motion
becomes large enough to emerge from the matching noise. The rule stops at the first clear fit rather
than selecting the strongest fit from a fixed temporal window, so an informative consecutive ping is
never delayed.

The pings between $i$ and $k$ retain their states in the factor graph but receive no F1 constraint for
this step; their motion is propagated locally through F0. Once $k$ is selected, it becomes the reference
ping of the next F1 step.

To prevent the interval from growing indefinitely, the step is limited to $\Delta t_{\max}$. If no
candidate reaches $\xi_{\min}$ within this interval, the fit with the largest significance is retained
with an inflated covariance. Its uncertainty is scaled according to the fraction of its inliers that
cannot be explained by chance,
\begin{equation}
\mat{\Sigma}_{z}'=\min\Big\{\frac{n}{n-\lambda_{ik'}},\,c_{\max}\Big\}\,\mat{\Sigma}_{z} ,
\label{eq:weakcov}
\end{equation}
where $c_{\max}$ limits the maximum inflation. A fit that returns no model is handled as a fit below
$\xi_{\min}$; if no usable fit can be retained, the corresponding interval is constrained only by F0.

\textbf{F1 factor.}
After the endpoint $k$ has been selected, the fitted planar motion and its covariance define the F1
measurement between the two keyframes. The factor compares this measurement with the relative pose
predicted by their graph states~\cite{dellaert2017factor},
\begin{equation}
\vect{e}_{\mathrm{F1}}=\Log\!\big(\meas{\mat{Z}}_{ik}^{-1}\,\mat{X}_{i}^{-1}\mat{X}_k\big) ,
\label{eq:f1}
\end{equation}
where $\meas{\mat{Z}}_{ik}\in\SE{3}$ is the planar motion $\meas{\vect{z}}_{ik}$ with zero roll, pitch
and heave. The factor uses $\mat{\Sigma}_z$ in the horizontal and yaw components, a loose covariance in
the remaining components, and a Huber kernel.

The metric character of F1 comes from the ground projection preceding the registration: each detected
feature is placed on the seabed using the altitude estimated from the seabed edge before its
displacement is measured. This projection assumes that the altitude measured near the seabed edge is
representative of the seabed under the features used in the step. A real seabed is not perfectly flat,
so its depth can vary across the footprint of a ping and the resulting projected points can differ from
their true positions. The next subsection quantifies how this departure from the local-flatness
assumption propagates into the error of the F1 motion estimate.

\subsection{Dependence of the step error on seabed geometry}\label{sec:f1-error}

We characterise the relative error $\eta=(\hat d-d)/d$ of a short, predominantly forward F1 step. The analysis first propagates an altitude error through feature projection and motion estimation, then relates that error to the seabed sampled by the two keyframes. We use a local model with fixed sonar orientation and explicitly identify the approximations required to neglect vertical motion.

\textbf{The projection error.}
Let $\delta h=\hat h-h$ be the error in the altitude used to project a feature. At a single ping, the measured slant range $r$ and bearing $\beta$ remain fixed while the inferred elevation $\psi$ changes. The corresponding altitude and along-track ground distance are
\begin{equation}
\begin{aligned}
h(\psi) &= r\big(\cos\psi\cos\beta\sin\alpha-\sin\psi\cos\alpha\big),\\
x(\psi) &= r\big(\cos\psi\cos\beta\cos\alpha+\sin\psi\sin\alpha\big).
\end{aligned}
\label{eq:hxpsi}
\end{equation}
Differentiating both quantities with respect to the same elevation gives
\begin{equation}
\begin{aligned}
\frac{\partial h}{\partial\psi}
&=-r\,(\cos\psi\cos\alpha+\sin\psi\cos\beta\sin\alpha),\\
\frac{\partial x}{\partial\psi}
&=r\,(\cos\psi\sin\alpha-\sin\psi\cos\beta\cos\alpha).
\end{aligned}
\label{eq:projection-derivatives}
\end{equation}
Where $\partial h/\partial\psi\ne0$, their ratio is $\partial x/\partial h$. A first-order Taylor expansion in $\delta h$ therefore yields
\begin{equation}
\begin{aligned}
\delta x&=\hat x-x\;\approx\;-\mu\,\delta h,\\
\mu&=\frac{\sin\alpha-\tan\psi\cos\beta\cos\alpha}
{\cos\alpha+\tan\psi\cos\beta\sin\alpha}.
\end{aligned}
\label{eq:xerr}
\end{equation}
The \emph{projection slope} $\mu$ is the along-track sensitivity to altitude error. Defining
\begin{equation}
\tan\psi'=\tan\psi\,\cos\beta ,
\label{eq:psiprime}
\end{equation}
gives
\begin{equation}
\mu=\tan(\alpha-\psi') .
\label{eq:mu}
\end{equation}
In the centre beam, $\psi'=\psi$ and $\mu=h/g$. For the forward-looking, downward-facing geometry considered here, with $0<\alpha-\psi_e<\pi/2$, the visible seabed features satisfy $\mu\ge0$ and $\psi'\ge\psi_e$, giving
\begin{equation}
\mu\;\le\;\tan(\alpha-\psi_e) .
\label{eq:slope-bound}
\end{equation}

\textbf{The relative step error.}
Consider a static feature $m$ observed in keyframes $i$ and $k$, separated by a forward displacement $d>0$. Let $h_{p,m}$ be the true sonar altitude over this feature in keyframe $p$, and define $\delta h_{p,m}=\hat h_p-h_{p,m}$, for $p\in\{i,k\}$. Equation~\eqref{eq:xerr} gives
\begin{equation}
\delta x_{i,m}\approx-\mu_{i,m}\,\delta h_{i,m},
\qquad
\delta x_{k,m}\approx-\mu_{k,m}\,\delta h_{k,m}.
\label{eq:step-proj}
\end{equation}
Since the true along-track coordinates differ by $d$, the feature's geometric contribution to the step error is
\begin{equation}
\delta d_m
=\delta x_{i,m}-\delta x_{k,m}
\approx
-\mu_{i,m}\delta h_{i,m}
+\mu_{k,m}\delta h_{k,m}.
\label{eq:step-feature}
\end{equation}
The change in altitude error is
\begin{equation}
\begin{aligned}
\Delta h
&=\delta h_{k,m}-\delta h_{i,m}\\
&=(\hat h_k-\hat h_i)-(h_{k,m}-h_{i,m}).
\end{aligned}
\label{eq:dh}
\end{equation}
It is common to the static matched features under a common vertical displacement, but equals $\hat h_k-\hat h_i$ only when their true altitudes are unchanged.

Let $\mu_m=\mu_{i,m}$ and $\mu'_m=(\partial\mu/\partial x)|_{x_m}$, with true altitude, cross-track position and sonar orientation held fixed. Because $x_{k,m}=x_m-d$, neglecting the contribution of vertical motion gives
\begin{equation}
\mu_{k,m}\approx\mu_m-d\,\mu'_m .
\label{eq:mu-step}
\end{equation}
If the vehicle changes height, the expansion also contains
$(h_{k,m}-h_{i,m})(\partial\mu/\partial h)_m$.
Equation~\eqref{eq:mu-step} is used only where that contribution is negligible at the retained accuracy.

Substituting $\delta h_{k,m}=\delta h_{i,m}+\Delta h$ and \eqref{eq:mu-step} into \eqref{eq:step-feature} produces an additional term $-d\,\mu'_m\Delta h$. For a terrain-induced error change $\Delta h=O(d)$, this term is second order in step length, leaving
\begin{equation}
\delta d_m
\approx
-d\,\mu'_m\,\delta h_{i,m}
+\mu_m\,\Delta h .
\label{eq:step-feature-first}
\end{equation}

For this scalar analysis, we approximate the fitted forward displacement by an equally weighted translation fit with rotation fixed. Its geometric error is then the mean contribution of the $n$ inliers. Let $\nu$ be the relative registration error due to feature localisation and matching, and define
\begin{equation}
\bar\mu=\frac{1}{n}\sum_m\mu_m .
\label{eq:mu-mean}
\end{equation}
The total step error $\delta d=\hat d-d$ is
\begin{equation}
\delta d
\approx
-\frac{d}{n}\sum_m\mu'_m\,\delta h_{i,m}
+\bar\mu\,\Delta h
+d\,\nu .
\label{eq:step-error}
\end{equation}
Dividing by $d$ gives
\begin{equation}
\eta\;\approx\;
-\frac1n\sum_m\mu'_m\,\delta h_{i,m}
+\bar\mu\,\frac{\Delta h}{d}
+\nu .
\label{eq:step-mean}
\end{equation}
Since $\hat d=(1+\eta)d$, $\eta$ is the relative scale error. The first two terms describe the effect of the altitude error in the reference keyframe and its change between keyframes; the last is the relative registration error. A joint rotation--translation fit additionally requires the along-track effect of rotation error to be negligible for this scalar reduction.

\textbf{The seabed model.}
Over the two overlapping footprints, we represent the same static seabed by a common inclined plane and a residual. In the coordinates of ping $k$,
\begin{equation}
\zeta_k(x)=\bar\zeta_k-(x-\bar x_k)\tan\tau+\epsilon_k(x) .
\label{eq:seabed}
\end{equation}
Here $\bar x_k$ is the footprint centre, $\bar\zeta_k$ the depth of the mean plane there, and $\tau$ the along-track tilt, positive when the seabed rises ahead. The residual $\epsilon_k(x)$ has depth spread $\sigma_\zeta$; its slope $\gamma_k(x)=-\mathrm{d}\epsilon_k/\mathrm{d}x$ has slope spread $\sigma_\gamma$. The subscript changes the sensor-relative coordinates, not the underlying seabed.

\textbf{The altitude error within one ping.}
Under the edge-ray model, and neglecting independent edge-detection error, the estimated altitude is the depth at the seabed edge,
\begin{equation}
\hat h_k=\zeta_k(g_e)=\bar\zeta_k-(g_e-\bar x_k)\tan\tau+\epsilon_k(g_e) .
\label{eq:hmeas}
\end{equation}
To isolate the error caused by using this edge depth for the matched features, we represent their depths by the local plane,
\begin{equation}
h_m\approx\bar\zeta_k-(x_m-\bar x_k)\tan\tau .
\label{eq:htrue}
\end{equation}
Here $h_m$ denotes $h_{k,m}$ within ping $k$. This is a modelling approximation: it neglects the residual $\epsilon_k(x_m)$ beneath each feature, and requires its contribution to the fitted step to be negligible. Subtraction gives
\begin{equation}
\delta h_{k,m}=\hat h_k-h_m\approx(x_m-g_e)\tan\tau+\epsilon_k(g_e) .
\label{eq:dHk}
\end{equation}

\textbf{The change between the two keyframes.}
The vehicle's vertical position cancels from the difference between the edge depth and a feature depth, but it still affects where the edge ray intersects the seabed. For a fixed edge-ray inclination,
$g_{e,k}=\hat h_k\cot(\alpha-\psi_e)$.
The physical displacement of the edge intersection is therefore
\begin{equation}
d_e=d+(\hat h_k-\hat h_i)\cot(\alpha-\psi_e) ,
\label{eq:edge-step}
\end{equation}
whereas the same matched feature becomes $d$ closer. Subtracting \eqref{eq:dHk} between keyframes consequently gives
\begin{equation}
\Delta h\approx-d_e\tan\tau
+\epsilon_k(g_{e,k})-\epsilon_i(g_{e,i}).
\label{eq:var-inc}
\end{equation}

The residual difference is a spatial change of the same seabed. Parameterising the interval between edge intersections by distance $\ell$, with $\ell=0$ at the first edge, gives
\begin{equation}
\epsilon_k(g_{e,k})-\epsilon_i(g_{e,i})
=-\int_0^{d_e}\gamma(\ell)\,\mathrm{d}\ell
\approx-d_e\,\gamma_k .
\label{eq:edge-slope-average}
\end{equation}
The approximation replaces the interval-average residual slope by the slope at the second edge, $\gamma_k=\gamma(d_e)$. If the residual is twice differentiable, the omitted term has magnitude at most
$\tfrac12 d_e^2\sup_{[0,d_e]}|\epsilon''|$.
Thus, it requires sufficiently small slope variation over the edge displacement.

Substitution into \eqref{eq:var-inc} gives
\begin{equation}
\Delta h\;\approx\;-d\,\Big[1+\frac{\hat h_k-\hat h_i}{d}\cot(\alpha-\psi_e)\Big]\big(\tan\tau+\gamma_k\big) .
\label{eq:dh-slope}
\end{equation}
We replace the bracket by one only when
\begin{equation}
\left|\frac{\hat h_k-\hat h_i}{d}\cot(\alpha-\psi_e)\right|\ll1 .
\label{eq:edge-small-change}
\end{equation}
This condition controls edge displacement. It is separate from neglecting the vertical-motion contribution to $\mu$ in \eqref{eq:mu-step}.

\textbf{Characterisation of the relative error.}
Substituting \eqref{eq:dHk} and the reduced form
$\Delta h/d\approx-(\tan\tau+\gamma_k)$
into \eqref{eq:step-mean} gives
\begin{equation}
\eta\approx-\mu_\tau\tan\tau
-\overline{\mu'}\,\epsilon_i(g_e)-\bar\mu\,\gamma_k+\nu ,
\label{eq:terrain-step}
\end{equation}
where $\overline{\mu'}=\frac1n\sum_m\mu'_m$ and
\begin{equation}
\mu_\tau=\frac1n\sum_m\big[\mu_m+(x_m-g_e)\,\mu'_m\big] .
\label{eq:tilt-gain}
\end{equation}
The mean tilt sets the systematic term; the edge-height residual, its slope and registration error supply the fluctuations.

Conditioned on the nominal feature geometry, assume that the sampled residual, its slope and $\nu$ have zero mean, and that $\nu$ is independent of the seabed. Before any further reduction, the variance is
\begin{equation}
\begin{aligned}
\sigma_\eta^2\approx\;&\sigma_\nu^2
+\overline{\mu'}^{\,2}\operatorname{Var}[\epsilon_i(g_e)]
+\bar\mu^{\,2}\sigma_\gamma^2\\
&+2\overline{\mu'}\,\bar\mu\,
\operatorname{Cov}\!\big(\epsilon_i(g_e),\gamma_k\big).
\end{aligned}
\label{eq:variance-full}
\end{equation}
For a reduced model dominated by the slope fluctuation, we neglect the edge-height fluctuation, requiring
$|\overline{\mu'}|\sqrt{\operatorname{Var}[\epsilon_i(g_e)]}
\ll|\bar\mu|\sigma_\gamma$.
The following bound applies to that reduced model.

\textbf{Proposition 2} (reduced local error model)\textbf{.}
Under the preceding geometric and statistical assumptions, and for feature and sonar configurations satisfying
$|\mu_\tau|\le\tan(\alpha-\psi_e)$,
the relative step error has the first-order mean
\begin{equation}
\mathrm{E}[\eta]\;\approx\;-\mu_\tau\tan\tau,\qquad
\big|\mathrm{E}[\eta]\big|\;\le\;\tan(\alpha-\psi_e)\,|\tan\tau| ,
\label{eq:scale-slope}
\end{equation}
and the variance of the reduced model satisfies
\begin{equation}
\sigma_\eta^2\;\le\;\sigma_\nu^2+\tan^2(\alpha-\psi_e)\,\sigma_\gamma^2 .
\label{eq:bound-seabed}
\end{equation}

\emph{Proof.} The mean is the expectation of \eqref{eq:terrain-step}. In the centre beam $\mu_m=h/x_m$ and $\mu'_m=-h/x_m^2$, so the term of feature $m$ in \eqref{eq:tilt-gain} is
\begin{equation}
\mu_m+(x_m-g_e)\,\mu'_m\;=\;\frac{h\,g_e}{x_m^2}\;\le\;\frac{h}{g_e}\;=\;\tan(\alpha-\psi_e) ,
\label{eq:proof-coef}
\end{equation}
since $x_m\ge g_e$. Evaluated over the rest of the beam for each sonar used in this work, its magnitude is largest at the seabed edge of the centre beam, so the condition on $\mu_\tau$ holds. Without the edge-height fluctuation, \eqref{eq:variance-full} reduces to
\begin{equation}
\sigma_\eta^2\;\approx\;\sigma_\nu^2+\bar\mu^{\,2}\,\sigma_\gamma^2 ,
\label{eq:bound-step}
\end{equation}
and \eqref{eq:slope-bound}, which also holds for the mean $\bar\mu$, gives \eqref{eq:bound-seabed}. $\square$

\subsection{Loop-closure factor (F2)}\label{sec:f2}

\textbf{Map construction.}
The \emph{seabed edge} used by F1 marks the onset of the bottom return. For mapping, we additionally select a nearby amplitude peak, termed the \emph{leading edge}. Its range $r_w(\beta)$ is selected within a short window around $r_e(\beta)$, jointly across beams by a dynamic program that balances amplitude against range jumps between adjacent beams.

We assign this peak range to the lower-edge ray to construct the 3-D point estimate
\begin{equation}
\vect{q}_k(\beta)=r_w(\beta)\,\mat{R}_k\,\vect{u}(\beta,\psi_e)+\vect{t}_{BS} ,
\label{eq:window}
\end{equation}
where $\mat{R}_k$ applies the mounting and roll--pitch rotations, with the lever arm expressed in the same point frame. Assigning the peak to $\psi_e$ is a geometric approximation: its elevation is not measured by the sonar. The pose-transformed points form a seabed map rasterised as a height grid $D(\cdot)$; the same poses place the flat-fielded images of \eqref{eq:flatfield} into a sonar mosaic.

\textbf{Map registration.}
The mosaic and seabed submaps provide two representations for detecting revisits. Both searches compare current and earlier observations using masked normalised cross-correlation~\cite{padfield2012masked}. For a map raster $M$ and a raster $P$ transformed by $(\vect{t},\theta)$,
\begin{equation}
\rho(\vect{t},\theta)=\frac{\sum_{\mathcal{O}}\breve{M}\,\breve{P}_{\vect{t},\theta}}{\sqrt{\sum_{\mathcal{O}}\breve{M}^2\,\sum_{\mathcal{O}}\breve{P}_{\vect{t},\theta}^2}} ,
\label{eq:ncc}
\end{equation}
where $\mathcal{O}$ is the overlap of their masks, and $\breve{M}$ and $\breve{P}$ are centred using their respective means over this overlap. For each candidate rotation, FFTs evaluate the translations simultaneously. The searches differ in their map representation and in the region examined around the current pose estimate.

\textbf{Mosaic closure.} A \emph{patch} $\mathcal{P}_j$ of the sonar mosaic, the last frames placed
with the current estimate, is compared with the mosaic $\mathcal{M}_i$ of the frames around a keyframe
$i$ of an earlier pass that the estimate places nearby and at a similar heading, over a narrow grid of
yaws around the odometry. The peak of \eqref{eq:ncc} is a candidate closure; it is accepted when its
peak, its margin to the best peak outside its neighbourhood and its overlap are large enough and when
the placements whose correlation comes within a margin of the peak are not spread out, which rejects
matches along an elongated band. Accepted candidates pass pairwise consistency
maximisation~\cite{mangelson2018pcm}, as in multi-robot SLAM~\cite{lajoie2020doorslam}, and a group
test that keeps a revisit only when several of its closures imply the same correction.

\textbf{Submap closure.} The mosaic closure needs the vehicle to come back close to where the estimate
expects it and at a similar heading. After a long detour neither holds: the heading drift of the odometry
can reach tens of degrees, and a survey vehicle often crosses its old track at a large angle. The submap
closure searches that case with the depth and the intensity of the seabed together. A query
$\mathcal{Q}_j$ holds the leading-edge points \eqref{eq:window} and the mosaic of the frames of the last
$T_q$ seconds, in the body frame of its middle frame $j$; a map submap $\mathcal{S}_i$ holds those of
the $T_m$ seconds around an earlier frame $i$, in the frame of $i$. Each is rasterised into a depth grid
$D$ and an intensity grid $I$, and both grids are band-passed by subtracting a wide Gaussian blur taken
over the valid cells only, which removes the tide and the slow trend of the echo level and keeps the
relief and the texture. For every yaw on a grid around the predicted relative pose
$\bar{\mat{X}}_i^{-1}\bar{\mat{X}}_j$, where $\bar{\mat{X}}_k\in\SE{2}$ is the planar part of
$\mat{X}_k$, on a window that widens with the time since frame $i$ to cover the heading drift, the two
correlations are added with the weight $w_I$ on the intensity,
\begin{equation}
\rho_{DI}(\vect{t},\theta)=\rho_D(\vect{t},\theta)+w_I\,\rho_I(\vect{t},\theta) ,
\label{eq:subscore}
\end{equation}
and the peak is taken over the placements whose overlap holds a minimum number of cells in both grids,
whose translation lies within a radius that grows with the distance travelled since frame $i$, and whose
mean depth difference over the overlap,
\begin{equation}
\Delta\bar z(\vect{t},\theta)=\frac{1}{|\mathcal{O}|}\sum_{\mathcal{O}}\big(D_{\mathcal{Q},\vect{t},\theta}-D_{\mathcal{S}}\big) ,
\label{eq:dz}
\end{equation}
is small. Below a surface vessel the depth is absolute up to the tide, so two places that look alike at
different depths are told apart. The depth fixes the position across the seabed relief and the heading,
the texture the position along it. The peak is refined on a grid of half the cell and accepted when its
score, its margin to the best placement farther away and the spread of the yaws whose score comes within
a margin of the peak pass their thresholds, and when the depth alone correlates by at least
$\rho_{D,\min}$ at the peak. Two passes over the same seabed see the same relief from any direction,
while the echo intensity changes with the direction of view, so a peak carried by the intensity alone is
more often a different place that looks alike than the same place seen again; in our implementation
$\rho_{D,\min}=0.7$. A registration $\mat{Z}_a$ of the query $j_a$ against the
map $i_a$ becomes a closure only when the registration $\mat{Z}_b$ of a later query $j_b$ against the
same region agrees with it through the odometry,
\begin{equation}
\big(\bar{\mat{X}}_{i_a}\mat{Z}_a\big)^{-1}\bar{\mat{X}}_{i_b}\mat{Z}_b\approx\bar{\mat{X}}_{j_a}^{-1}\bar{\mat{X}}_{j_b} ,
\label{eq:agree}
\end{equation}
within a tolerance in position and heading: the two closures imply the same relative pose of the two
queries as the odometry between them. Two confident peaks at the wrong places are rarely consistent in
this way.

Both kinds enter the graph as planar relative poses between the two body frames, with a Cauchy kernel;
the covariance of a mosaic closure is set by the resolution of its correlation grid, that of a submap
closure by its raster cell in position and by the turn inside the query in heading. A submap closure can
ask for a correction of metres and tens of degrees, far in the tail of the Cauchy kernel at the odometry
estimate, so the graph is solved first with the closure covariances inflated and then with the nominal
ones.

\subsection{The graph}\label{sec:map}
With $\mathcal{X}=\{\mat{X}_k,\vect{v}_k,\vect{b}_k\}$ at every ping, the estimate solves
\begin{equation}
\begin{aligned}
\mathcal{X}^\star=\arg\min_{\mathcal{X}}\ &\textstyle\sum\|\vect{e}_{\mathrm{F0}}\|^2+\sum\rho_{\mathrm{H}}\big(\|\vect{e}_{\mathrm{F1}}\|^2\big)\\
&+\textstyle\sum\rho_{\mathrm{C}}\big(\|\vect{e}_{\mathrm{F2}}\|^2\big)+\|\vect{e}_{\mathrm{prior}}\|^2 ,
\end{aligned}
\label{eq:map}
\end{equation}
where each norm is weighted by the covariance of its factor, $\rho_{\mathrm{H}}$ and $\rho_{\mathrm{C}}$
are the Huber and Cauchy losses, and $\vect{e}_{\mathrm{prior}}$ holds the first-pose prior, the bias
priors and the vertical prior \eqref{eq:heave}. The problem is solved by Levenberg--Marquardt in
GTSAM~\cite{gtsam43}, incrementally, and re-solved after new closures, so that later searches start
from a corrected estimate.

Only F1 measures a metric translation in \eqref{eq:map}, and this is where the scale of
\cref{sec:sonar-scale} is settled. By \eqref{eq:step-mean} an altitude error rescales the fitted
translation of a step by $s=1+\eta$ and leaves its rotation unchanged to first order, so every F1 factor
pulls its step to that scale. The other factors do not change it: by Proposition~1, on the legs of a
survey a step scaled by $s$ is matched by a shifted accelerometer bias and $\vect{e}_{\mathrm{F0}}$ does
not change, and the loop closures compare maps built with the same projection, so a trajectory scaled by
a constant still closes every loop. The minimum of \eqref{eq:map} therefore keeps the scale of F1, and
\brina{} fixes it by measurement: the altitude is an input of F1, not a state of the graph, and its
error is the one of \cref{sec:f1-error}.

\section{Experiments and results}\label{sec:exp}
The experiments evaluate whether the altitude estimated from the seabed edge provides metric scale to sonar odometry, and whether the remaining relative step error follows the model in \cref{sec:f1-error}. \Cref{sec:sim} describes the simulator and the seabed-following AUV mission, then uses the known seabed geometry and ground-truth poses to evaluate the predictions and bounds of Proposition~2 for the bias and spread of the step error across straight legs and turns. \Cref{sec:field} describes the USV platforms and sonar configurations, then reports trajectory accuracy in calm water and waves. For a mission with repeated revisits, we compare the position error with and without loop closures over the course of the trajectory.

\subsection{Simulations}\label{sec:sim}
\textbf{Simulator and seabed-following mission.}
Stonefish~\cite{cieslak2019stonefish} is used to simulate an AUV carrying a forward-looking sonar with the geometry of the 900~kHz BlueView used in the field experiments. The AUV follows the same 400~m survey at 2~m/s over ten simulated seabeds. Each seabed combines two independently generated depth variations: one with a 50~m correlation length that determines the tilt across a sonar footprint, and one with a 4~m correlation length that introduces local roughness.

The survey consists of four parallel legs spaced 20~m apart, joined by semicircular turns of 10~m radius. A proportional depth controller makes the AUV follow the seabed profile while targeting an altitude of 20~m above the seabed at the seabed edge. To generate the trajectory, the controller uses the altitude computed from the known seabed at the edge, with added bias and noise to model the edge detector. Stonefish then renders the sonar pings along this trajectory. \brina{} processes the resulting sonar and IMU data, and the known vehicle poses provide the reference displacement for each F1 step.

\textbf{Evaluation of Proposition~2.}
The bias and spread predicted by Proposition~2 were evaluated over
10\,034 F1 steps from the ten simulated missions, including both
straight legs and turns. For each step, the relative error was computed
as $\eta=(\hat d-d)/d$, with the reference displacement $d$ obtained
from the ground-truth poses. In \cref{tab:simmatrix}, the steps are grouped by
the along-track tilt $\tau$ of the seabed footprint and the local slope
spread $\sigma_\gamma$. For each reported class, the measured mean and standard
deviation are shown alongside their model estimates. The matching-noise standard deviation, $\sigma_\nu=3.6\%$,
was estimated from the smooth, near-level class and used throughout.

\begin{table}[t]
\caption{Proposition~2 on the 10\,034 F1 steps of the ten AUV missions, straight legs and turns together, by class of
tilt $\tau$ of the footprint (rows) and of roughness $\sigma_\gamma$ (columns). All values in percent. In each box, top, on blue:
measured, sample mean $\pm$ sample standard deviation of the relative step error $\eta$; middle, on orange: predicted,
mean \eqref{eq:scale-slope} $\pm$ standard deviation \eqref{eq:bound-step}; bottom, in brackets: the upper bounds of
Proposition~2, on $|\mathrm{E}[\eta]|$ \eqref{eq:scale-slope} and on $\sigma_\eta$ \eqref{eq:bound-seabed}. Every standard
deviation is that of $\eta$ less the predicted mean of its own step, so the change of the mean inside a class does not enter
it; the predicted one and its bound are root mean squares of the per-step values, and the bound on the mean is the class
average of the per-step bounds $\tan(\alpha-\psi_e)|\tan\tau|$. $\sigma_\nu=3.6\%$, measured on the smooth and level box.
Both bounds hold in all thirteen boxes. A dash: fewer than 40 steps in the box.}
\label{tab:simmatrix}
\centering
\footnotesize
\renewcommand{\arraystretch}{1.0}
\setlength{\tabcolsep}{5pt}
\begin{NiceTabular}{|l|c|c|c|}
\CodeBefore
  \cellcolor{measc}{3-3,3-4,6-2,6-3,6-4,9-2,9-3,9-4,12-2,12-3,12-4,15-2,15-4}
  \cellcolor{predc}{4-3,4-4,7-2,7-3,7-4,10-2,10-3,10-4,13-2,13-3,13-4,16-2,16-4}
\Body
\Hline
\Block[l]{2-1}{tilt of the footprint} & \Block{1-3}{roughness $\sigma_\gamma$} & & \\ \cline{2-4}
 & $<2^\circ$ & $2^\circ$ to $5^\circ$ & $>5^\circ$ \\ \Hline
\Block[l]{3-1}{falling, $\tau<-6^\circ$} & \Block{3-1}{--} & $+5.7$ $\pm$ 4.8 & $+7.6$ $\pm$ 10.2 \\
 &  & $+7.2$ $\pm$ 5.4 & $+6.8$ $\pm$ 11.3 \\
 &  & (10.4,\,6.1) & (9.9,\,13.8) \\ \Hline
\Block[l]{3-1}{$-6^\circ$ to $-2^\circ$} & $+3.1$ $\pm$ 1.3 & $+3.9$ $\pm$ 4.1 & $+3.8$ $\pm$ 11.0 \\
 & $+2.3$ $\pm$ 3.8 & $+3.1$ $\pm$ 5.0 & $+3.4$ $\pm$ 10.2 \\
 & (3.6,\,4.0) & (4.9,\,5.8) & (5.2,\,12.6) \\ \Hline
\Block[l]{3-1}{level, $|\tau|<2^\circ$} & $+0.2$ $\pm$ 3.6 & $+0.3$ $\pm$ 4.4 & $+0.0$ $\pm$ 10.9 \\
 & ${-}0.3$ $\pm$ 3.8 & $+0.0$ $\pm$ 4.9 & $+0.0$ $\pm$ 8.8 \\
 & (0.9,\,4.0) & (1.2,\,5.7) & (1.2,\,11.4) \\ \Hline
\Block[l]{3-1}{$2^\circ$ to $6^\circ$} & ${-}1.0$ $\pm$ 2.7 & ${-}2.6$ $\pm$ 4.0 & ${-}2.8$ $\pm$ 9.7 \\
 & ${-}2.6$ $\pm$ 4.0 & ${-}2.8$ $\pm$ 4.8 & ${-}2.7$ $\pm$ 9.2 \\
 & (4.5,\,4.3) & (4.6,\,5.6) & (4.4,\,12.1) \\ \Hline
\Block[l]{3-1}{rising, $\tau>6^\circ$} & ${-}5.2$ $\pm$ 3.4 & \Block{3-1}{--} & ${-}4.3$ $\pm$ 6.2 \\
 & ${-}5.1$ $\pm$ 4.0 &  & ${-}5.4$ $\pm$ 10.5 \\
 & (8.6,\,4.3) &  & (8.9,\,13.7) \\ \Hline
\end{NiceTabular}
\end{table}

The dependence of the bias on seabed tilt is shown by comparing rows
within a column of \cref{tab:simmatrix}. At intermediate roughness, step length
was overestimated by $3.9\%$ on falling terrain with tilt magnitudes
between $2^\circ$ and $6^\circ$, and underestimated by $2.6\%$ on
rising terrain in the same range. The corresponding predicted mean
errors were $+3.1\%$ and $-2.8\%$. Across all reported classes, the
measured and predicted means differed by at most 1.6 percentage
points. In every tilted class, the magnitude of the measured mean
remained within the bias bound of \eqref{eq:scale-slope},
\[
  |\mathrm{E}[\eta]| \leq \tan(\alpha-\psi_e)\,|\tan\tau|.
\]

The dependence of the spread on seabed roughness is shown by comparing
columns within a row of \cref{tab:simmatrix}. On near-level terrain, standard
deviations of $3.6\%$, $4.4\%$, and $10.9\%$ were measured with
increasing roughness, while the magnitude of the mean error remained
at or below $0.3\%$. Steps taken during turns are included in these
statistics. As the vehicle turns, the seabed edge sweeps laterally
and depth variations across the track are also sampled. In three
rough-terrain classes, the measured standard deviation exceeded the
model estimate by up to 2.2 percentage points. In all 13 reported
classes, the measured standard deviation remained within the upper
bound obtained from \eqref{eq:bound-seabed},
\[
  \sigma_\eta \leq
  \sqrt{\sigma_\nu^2+\tan^2(\alpha-\psi_e)\,\sigma_\gamma^2}.
\]

\begin{figure}[t]
\centering
\includegraphics[width=\columnwidth]{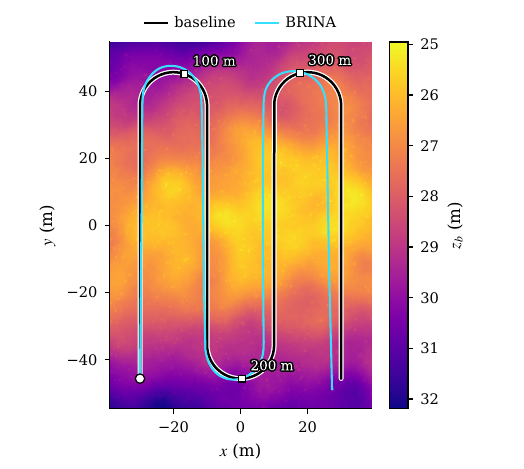}
\caption{Mission ID$_{8}$ from above, east $x$ and north $y$. The colour is the true depth $z_b$ of the
seabed. The black line is the true path, starting at the white circle, with a square every
\SI{100}{\metre}, and the cyan line the \brina{} trajectory, odometry only.}
\label{fig:auvmap}
\end{figure}

\begin{figure}[t]
\centering
\includegraphics[width=\columnwidth]{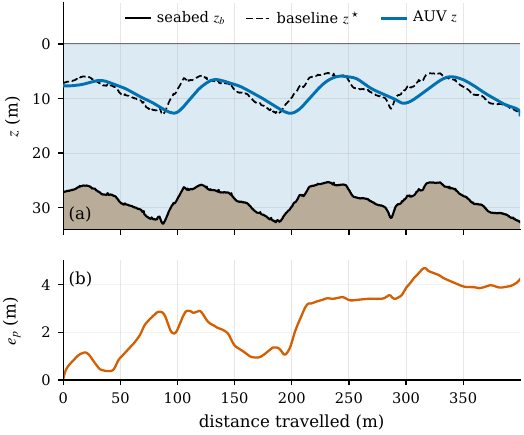}
\caption{Mission ID$_{8}$ against the distance travelled. (a) Depth $z$ of the sonar (blue), depth $z_b$
of the seabed where the centre edge ray meets it, about \SI{28}{\metre} ahead of the vehicle
(black), and target depth $z^\star=z_b-h^\star$ (dashed), \SI{20}{\metre} above the seabed.
(b) Position error $e_p$ of \brina{}.}
\label{fig:auvprofile}
\end{figure}

At the trajectory level, mean horizontal position errors between
1.30~m and 7.48~m were obtained over the ten 400~m surveys
(\cref{tab:sim}). Mission ID$_8$ is illustrated in \cref{fig:auvmap,fig:auvprofile}. In \cref{fig:auvmap},
the \brina{} trajectory and the ground-truth path are overlaid on the
seabed depth map, showing their separation along the survey legs and
through the turns. In \cref{fig:auvprofile}(a), the sonar depth is compared with the
target depth set 20~m above the seabed at the first return. The
horizontal position error is plotted against distance travelled in
\cref{fig:auvprofile}(b). A mean error of 2.55~m and a final error of 4.21~m were
obtained for this mission.

\begin{table}[t]
\caption{The ten simulated AUV missions, ID$_1$ to ID$_{10}$ in order of roughness. Tilt: root-mean-square tilt $\tau$ of the
footprint along the track. Roughness: slope spread $\sigma_\gamma$ of the seabed under the seabed edge, on the straight legs.
$e_p$: position error in the horizontal plane. $e_z$: error of the vertical position, which \brina{} measures from the seabed at
the seabed edge. Both are the mean and the standard deviation over the \SI{400}{\metre} of the mission.}
\label{tab:sim}
\centering
\footnotesize
\begin{tabular}{@{}lcccc@{}}
\toprule
ID & tilt ($^\circ$) & roughness $\sigma_\gamma$ ($^\circ$) & $e_p$ (m) & $e_z$ (m) \\
\midrule
ID$_{1}$ & 0.7 & 1.0 & 1.30 $\pm$ 0.70 & $+0.52$ $\pm$ 0.07 \\
ID$_{2}$ & 2.1 & 1.4 & 2.19 $\pm$ 1.20 & $+0.55$ $\pm$ 0.08 \\
ID$_{3}$ & 4.4 & 2.9 & 3.00 $\pm$ 1.75 & $+0.48$ $\pm$ 0.13 \\
ID$_{4}$ & 3.0 & 3.3 & 4.07 $\pm$ 2.33 & $+0.54$ $\pm$ 0.14 \\
ID$_{5}$ & 3.9 & 3.3 & 6.44 $\pm$ 4.25 & $+0.56$ $\pm$ 0.16 \\
ID$_{6}$ & 2.7 & 4.2 & 3.01 $\pm$ 1.83 & $+0.52$ $\pm$ 0.13 \\
ID$_{7}$ & 5.8 & 5.0 & 7.48 $\pm$ 5.49 & $+0.62$ $\pm$ 0.17 \\
ID$_{8}$ & 4.6 & 5.6 & 2.55 $\pm$ 1.38 & $+0.61$ $\pm$ 0.23 \\
ID$_{9}$ & 2.6 & 5.7 & 4.76 $\pm$ 2.83 & $+0.55$ $\pm$ 0.22 \\
ID$_{10}$ & 5.3 & 7.0 & 6.52 $\pm$ 3.58 & $+0.57$ $\pm$ 0.23 \\
\bottomrule
\end{tabular}
\end{table}

\subsection{Real-world experiments}\label{sec:field}

\textbf{Platforms and missions.}
\brina{} was evaluated on the ten USV missions listed in \cref{tab:missions}.
Missions A--I were recorded with a Maritime Robotics Otter equipped
with a bow-mounted forward-looking sonar and an Ouster OS2, whose
IMU provided the inertial measurements for \brina{}. A Teledyne BlueView
was operated at 900~kHz or 2.25~MHz on missions A--G, and an Oculus
was operated at 750~kHz on missions H and I. Mission J was recorded
in Vodice, Croatia, using a second USV equipped with an Oculus
operating at 1.2~MHz and its own IMU. The trials included calm-water
harbour surveys and two lake runs in waves (missions F and G).
Several harbour missions included repeated passes over the same
seabed, allowing the effect of loop closures to be assessed.

\begin{table}[t]
\centering
\caption{Field missions. Freq.: sonar frequency. Range: window of the sonar image.}
\label{tab:missions}
\footnotesize
\begin{tabular}{@{}lccc@{}}
\toprule
Mission & sonar & freq. (kHz) & range (\si{\metre}) \\
\midrule
A & BlueView & 900 & 1.2--8 \\
B & BlueView & 2250 & 1.2--8 \\
C & BlueView & 2250 & 1.2--8 \\
D & BlueView & 2250 & 1.2--10 \\
E & BlueView & 900 & 1.2--20 \\
F & BlueView & 900 & 1.2--20 \\
G & BlueView & 900 & 1.2--20 \\
H & Oculus & 750 & 0--10 \\
I & Oculus & 750 & 0--10 \\
J & Oculus & 1200 & 0--25 \\
\bottomrule
\end{tabular}
\end{table}

\brina{} and the reference trajectories were aligned in time and position
using the offsets between their first ROS~2 messages. Trajectory accuracy
was evaluated against GNSS or FAST-LIO2 LiDAR--inertial
odometry~\cite{xu2022fastlio2}.
Mean, root-mean-square, and final horizontal position errors were
computed relative to the reference trajectory. To quantify drift
over a fixed distance, the estimated and reference displacements
were also compared for pairs of poses separated by 100~m along the
reference trajectory. The magnitude of each displacement error was
expressed as a percentage of 100~m and averaged over all such pairs.

\begin{figure*}[t]
\centering
\includegraphics[width=\textwidth]{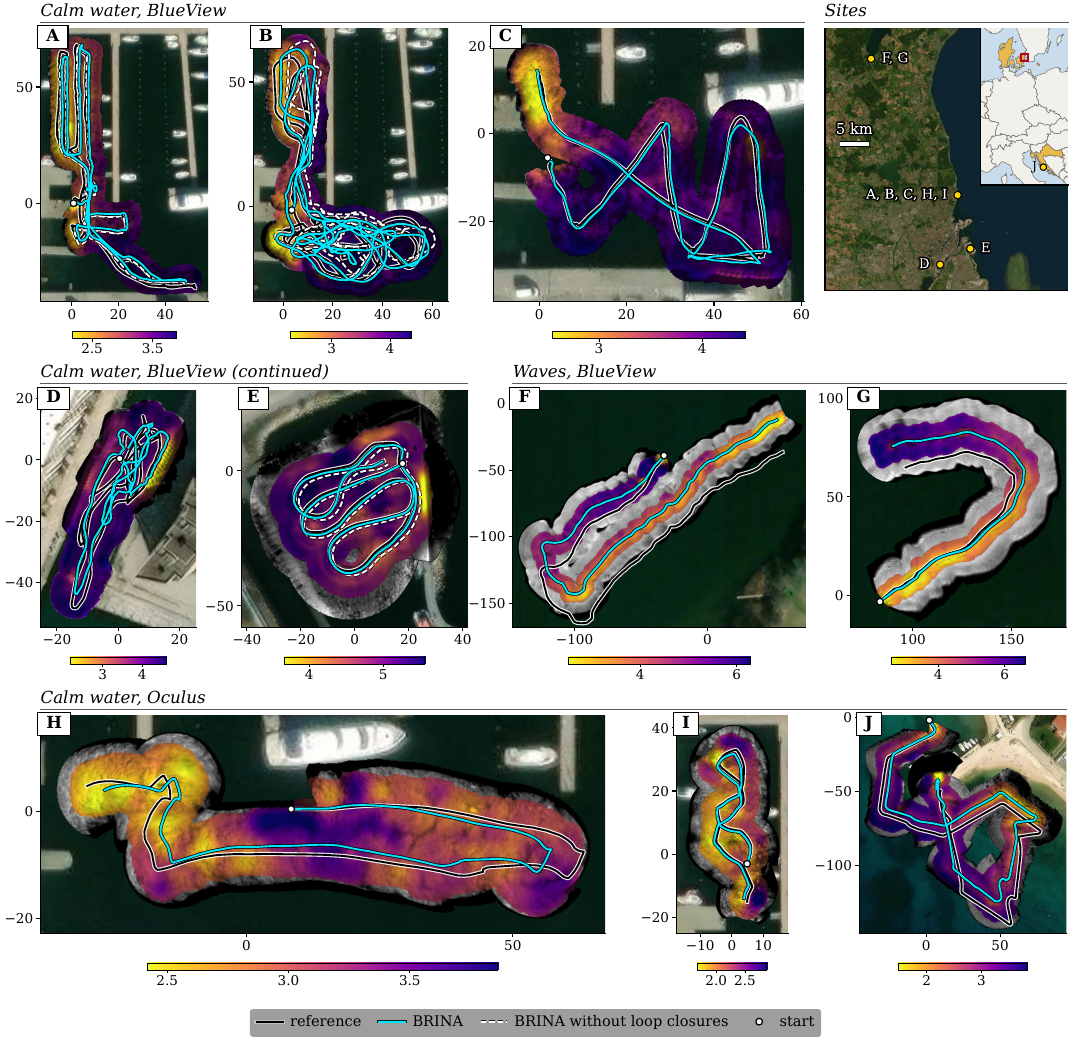}
\caption{The ten field missions (\cref{tab:missions}), grouped by water condition and sonar and
lettered from left to right; the panel at the top right shows where they were recorded, on the
satellite image of the Copenhagen area, and its inset marks that area and Vodice (J), in Croatia, on the
map of Europe. The black line is the reference track and the cyan line the \brina{} trajectory,
rotated by the rotation of the rigid fit of the whole trajectory onto the reference and translated so
that both start at the white circle. The grey image is the sonar mosaic placed with the same \brina{}
poses, over the satellite image of the site (Esri World Imagery); it is coloured by the depth of the
seabed below the sonar wherever a seabed-edge point of \cref{sec:f1} falls, with the scale in
metres under each panel. On A, B, D and E, the missions with accepted loop closures, the dashed white line
is \brina{} without them, matched in the same way; on D it stays within \SI{0.4}{\metre} of the cyan line. Mission B has no GNSS: it is placed on the satellite image by registering
its depth map to the depth maps of A and C. On D, H and I the LiDAR reference is turned to east and north by its rigid fit to
the GNSS fixes. Mission E has no GNSS either: its LiDAR reference is placed on the satellite image by
registering its LiDAR map, the Ouster scans placed with the FAST-LIO2 poses, to the quay edges of the
image. Axes in metres, east and north.}
\label{fig:fieldmaps}
\end{figure*}

\textbf{Field results.}
The estimated trajectories are compared with their references in
\cref{fig:fieldmaps}, and the corresponding errors are reported in \cref{tab:field}.
On the calm-water BlueView missions A and C, the estimated path
lengths were within 7\% of the reference lengths at both sonar
frequencies. Mean horizontal position errors of 2.20~m and 1.16~m
were obtained, respectively. These results support the use of the
seabed-edge altitude to recover metric scale in the field. Across
the five calm-water BlueView missions (A--E), the displacement
error over 100~m ranged from 0.7\% to 4.2\%. Larger errors of
7.7\% and 9.4\% were measured on the two lake runs in waves
(F and G), where greater roll and pitch variations were recorded.
With the Oculus, the displacement error over 100~m ranged from
1.2\% to 6.6\% across missions H--J.

\begin{table*}[t]
\centering
\caption{Field results by category, all with the loop closures of \brina{} on; A and B also without. Position error against the reference, GNSS on A, C, F, G and J and FAST-LIO2 LiDAR odometry on B, D, E, H and I, with the trajectory matched as in \cref{fig:fieldmaps}: mean, root mean square (RMSE) and standard deviation (std). Per \SI{100}{\metre}: error of the displacement over every \SI{100}{\metre} travelled, in total and split into its along-track part (signed; negative when \brina{} travels too little) and its cross-track part. Path: length of the \brina{} trajectory. Roll and pitch: mean $\pm$ standard deviation over the mission, from the IMU, in the IMU frame, so that the means include the mounting.}
\label{tab:field}
\footnotesize
\setlength{\tabcolsep}{5pt}
\begin{tabular}{@{}lccccccccc@{}}
\toprule
 & \multicolumn{3}{c}{position error (\si{\metre})} & \multicolumn{3}{c}{per \SI{100}{\metre} (\%)} & path & roll & pitch \\
\cmidrule(lr){2-4}\cmidrule(lr){5-7}
Mission & mean & RMSE & std & total & along & cross & (\si{\metre}) & (\si{\degree}) & (\si{\degree}) \\
\midrule
\multicolumn{10}{@{}l}{\emph{Loop closures}} \\
A, with LC & 2.20 & 2.51 & 1.20 & 3.1 & $-0.8$ & 1.8 & 458 & $-1.2 \pm 0.9$ & $-3.0 \pm 0.5$ \\
A, without LC & 2.25 & 2.62 & 1.34 & 3.1 & $-0.8$ & 1.7 & 457 &  & \\
B, with LC & 3.35 & 3.79 & 1.77 & 2.9 & $+1.4$ & 1.8 & 824 & $-1.1 \pm 0.9$ & $-3.2 \pm 0.4$ \\
B, without LC & 5.55 & 6.21 & 2.80 & 4.1 & $+1.2$ & 3.1 & 824 &  & \\
\midrule
\multicolumn{10}{@{}l}{\emph{Calm water, BlueView}} \\
C & 1.16 & 1.33 & 0.64 & 1.8 & $-0.2$ & 1.4 & 253 & $-1.0 \pm 0.9$ & $-3.1 \pm 0.6$ \\
D & 4.00 & 5.01 & 3.01 & 4.2 & $-1.3$ & 3.1 & 262 & $-1.2 \pm 0.8$ & $-2.7 \pm 0.4$ \\
E & 0.86 & 0.94 & 0.38 & 0.7 & $+0.2$ & 0.3 & 308 & $-1.0 \pm 0.9$ & $-2.9 \pm 0.4$ \\
\midrule
\multicolumn{10}{@{}l}{\emph{Waves}} \\
F & 17.97 & 18.79 & 5.47 & 7.7 & $-3.3$ & 4.9 & 378 & $-1.3 \pm 1.5$ & $-3.8 \pm 2.5$ \\
G & 5.52 & 6.98 & 4.27 & 9.4 & $+9.3$ & 1.4 & 181 & $-1.3 \pm 2.4$ & $-3.0 \pm 2.4$ \\
\midrule
\multicolumn{10}{@{}l}{\emph{Calm water, Oculus}} \\
H & 3.39 & 3.67 & 1.42 & 6.6 & $-6.6$ & 0.4 & 159 & $-0.1 \pm 0.4$ & $-2.3 \pm 0.5$ \\
I & 0.88 & 0.94 & 0.33 & 1.2 & $-0.8$ & 0.8 & 106 & $-0.2 \pm 0.2$ & $-2.2 \pm 0.4$ \\
J & 9.46 & 10.71 & 5.04 & 6.6 & $-4.8$ & 3.4 & 377 & $+0.9 \pm 2.7$ & $+0.9 \pm 2.2$ \\
\bottomrule
\end{tabular}
\end{table*}

\textbf{Effect of loop closures.}
Mission B was selected to examine how the trajectory was corrected
when previously surveyed areas were revisited. Over the 804~m
reference path, the same harbour corridor was traversed several
times and the track was crossed repeatedly in the adjoining basin
(\cref{fig:fieldmaps}). Loop-closure constraints were accepted for 49 pose pairs
from seabed-submap registration and 26 from sonar-mosaic registration.
Their combined effect was evaluated by comparing the trajectories
estimated with and without these constraints against the same
FAST-LIO2 reference.

\begin{figure*}[t]
\centering
\includegraphics[width=\textwidth]{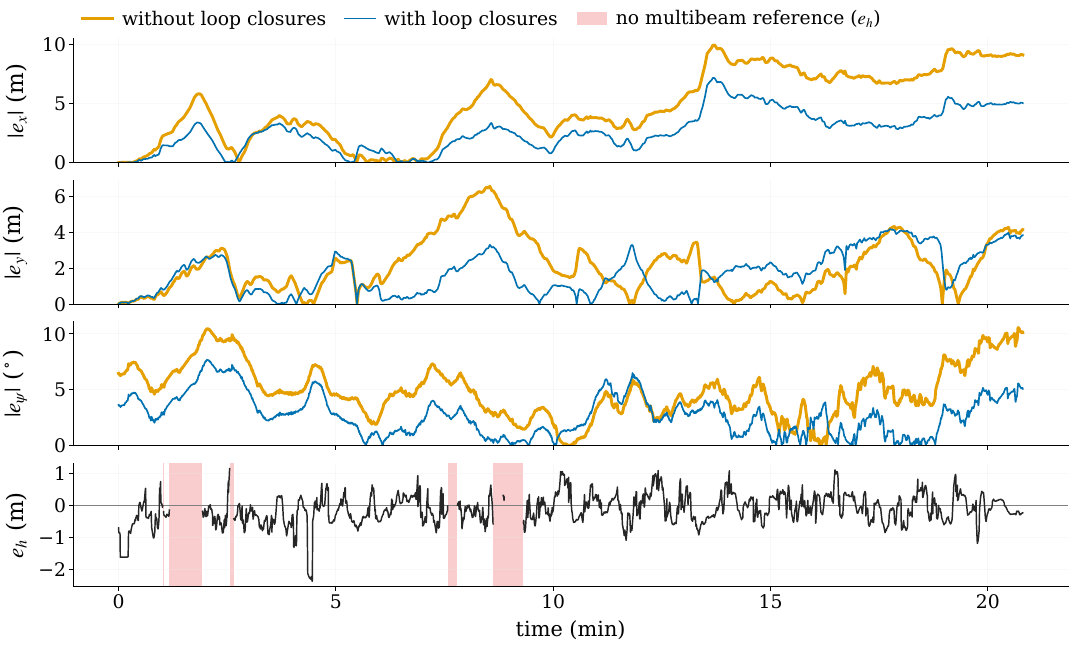}
\caption{Errors of \brina{} against time along mission B (\SI{21}{\minute}, \SI{804}{\metre}). From top to bottom: absolute east and north position error
$|e_x|$ and $|e_y|$ and absolute heading error $|e_\psi|$ against the FAST-LIO2 odometry, after the match
of \cref{fig:fieldmaps}; and the altitude error $e_h$, the seabed-edge altitude $\hat h_k$ of \eqref{eq:Hk}
minus the depth at which the centre edge ray meets the multibeam survey of the harbour, made the same day,
with the sonar at the depth of the multibeam head (the two heads sit at about the same height). The thick
orange line is \brina{} without loop closures and the thin blue line \brina{} with them; $e_h$ is the same
for both. The red bands mark where the centre edge ray falls outside the multibeam survey, on the east side
of the northern corridor: there is no multibeam reference there and $e_h$ is not computed.}
\label{fig:fielderr}
\end{figure*}

The evolution of the position error over mission B is shown in
\cref{fig:fielderr}. With loop closures, the mean error was reduced from 5.55~m
to 3.35~m, and the final error after 804~m was reduced from 10.0~m
to 6.3~m. The displacement error over 100~m was also reduced, from
4.1\% to 2.9\% (\cref{tab:field}). The largest improvement was observed in
the cross-track component, which decreased from 3.1\% to 1.8\%,
while the signed along-track component changed only slightly, from
$+1.2\%$ to $+1.4\%$. On mission A, the mean position error changed
only from 2.25~m to 2.20~m. The larger correction on mission B
shows how repeated observations of the same area reduced accumulated
drift over that survey.

\section{Conclusion}\label{sec:concl}
This work introduced \brina{}, a sonar--inertial SLAM method that recovers metric motion from a forward-looking sonar and an IMU, without a DVL. Its central contribution is to use the first seabed return in each beam as an altitude reference for odometry. The return range, beam geometry, and IMU attitude give the sonar altitude above the seabed. This altitude allows matched AKAZE features to be projected onto a local seabed plane, where the motion between pings is estimated in metres. A factor graph combines these motion estimates with IMU preintegration and loop closures from sonar mosaics and seabed submaps.

The analysis showed why the altitude reference is needed for the straight legs, constant accelerations, and steady turns considered here, during which the IMU alone cannot resolve the scale of sonar motion. Once altitude is available, the along-track tilt of the seabed within a ping determines the predicted bias in step length, and local slope variability bounds the terrain contribution to its spread. In the straight-leg model, successive changes in edge altitude sum to the difference between the two endpoints.

The error model was evaluated in simulation with a seabed-following AUV. Across 10\,034 steps over ten simulated seabeds, predicted and measured mean errors agreed to within 1.6 percentage points, and the measured spread remained within the analytical bound in all 13 terrain classes. Field performance was evaluated on ten USV missions at six sites with BlueView and Oculus sonars. Mean displacement error over 100~m ranged from 0.7\% to 4.2\% on the five calm-water BlueView surveys and reached 7.7\% and 9.4\% on the two runs in waves. On an 804~m harbour survey with repeated revisits, loop closures from seabed submaps and sonar mosaics together reduced mean horizontal position error from 5.55~m to 3.35~m.

By linking motion scale to an observable feature of the sonar image and its uncertainty to local seabed geometry, \brina{} provides a basis for sonar--inertial navigation wherever the seabed is visible to the sonar. Future work will take the seabed-following trials from simulation to field experiments with an AUV and extend \brina{} to collaborative navigation, in which multiple robots use a shared seabed map to estimate their poses.

\section*{Declaration of generative AI and AI-assisted technologies in the writing process}
The authors used Anthropic Claude to draft parts of the text and the scripts that produce
the tables and figures, and OpenAI GPT OSS to improve phrasing. They reviewed and edited the content and take
full responsibility for the publication.

\bibliographystyle{IEEEtran}
\bibliography{refs}
\end{document}